\documentclass[acmsmall,screen,nonacm]{acmart}

\setcopyright{none}
\usepackage{colortbl}
\usepackage{multirow}
\usepackage{amsthm}
\theoremstyle{definition}

\usepackage{graphicx} %
\usepackage{tabularx}
\usepackage{subcaption}
\usepackage{amsmath}
\usepackage{array}
\usepackage{graphicx}
\usepackage{subcaption}
\usepackage{xcolor}
\usepackage{tikz}
\usetikzlibrary{arrows.meta}

\usepackage{tcolorbox}

\newtcolorbox{resultbox}{
  colback=black!4,
  colframe=black!35,
  boxrule=0.4pt,
  arc=1mm,
  boxsep=0pt,
  left=6pt,
  right=6pt,
  top=5pt,
  bottom=5pt,
  before skip=6pt,
  after skip=8pt
}

\title{Counterfactual Model Testing Under Distribution Shift}

\author{Joran Leest}
\orcid{0009-0003-1335-6559}
\affiliation{%
  \institution{Vrije Universiteit Amsterdam}
  \country{The Netherlands}
}
\email{j.g.leest@vu.nl}

\author{Ilias Gerostathopoulos}
\orcid{0000-0001-9333-7101}
\affiliation{%
  \institution{Vrije Universiteit Amsterdam}
  \country{The Netherlands}
}
\email{i.g.gerostathopoulos@vu.nl}

\author{Claudia Raibulet}
\orcid{0000-0002-7194-3159}
\affiliation{%
  \institution{Universita' degli Studi di Milano-Bicocca}
  \country{Italy}
}
\email{claudia.raibulet@unimib.it}

\author{Patricia Lago}
\orcid{0000-0002-2234-0845}
\affiliation{%
  \institution{Vrije Universiteit Amsterdam}
  \country{The Netherlands}
}
\email{p.lago@vu.nl}

\begin{abstract}

Deployed machine learning (ML) models can degrade as the distribution of their input data shifts. Such shifts can result from changes in the population or from data integrity issues in the pipeline that materializes model features, two sources that operate through different mechanisms. Because labels are often absent or delayed, it is common practice to track shifts in the data instead, but a detected shift does not indicate how it affects the model. We formulate this problem as online model testing under distribution shift. Testing requires reproducing the shift on a controlled baseline with known labels. Since the same shift can come from either source, doing so requires knowing which mechanism changed. We propose a methodology that reproduces observed changes in both mechanisms to estimate current performance and the potential gain from repairing observed data integrity issues. Experiments on real-world datasets with simulated faults show that separating the two sources generally improves both estimates, supporting this formulation. A churn prediction case illustrates how these tests can support monitoring and repair prioritization.

\end{abstract}

\begin{document}

\maketitle

\section{Introduction}\label{sec:introduction}
Machine learning (ML) models provide software applications with capabilities such as product ranking, fraud detection, and churn prediction. Once trained, a model is deployed under the assumption that the distribution it encounters remains stationary \cite{liu2023need, huyen2022}. However, deployment exposes the model to two sources of change that break this assumption. The population (e.g., the mix of users and their behavior) changes continuously through events, trends, and seasonality \cite{vzliobaite2015overview, leest2025monitoring}, and data integrity issues are introduced as the pipelines producing the model's features change, break, or become stale \cite{shankars2024we, cavenesse2020tensorflow, leest2025monitoring, shankar2023automatic}. Both introduce distribution shifts that can degrade performance until predictions are no longer reliable for the decisions they support. Deployment therefore begins an ongoing process of monitoring and response \cite{shankars2024we,leest2025monitoring}.

In practice, engineers track feature and data integrity metrics on dashboards and follow on-call processes to investigate alerts and respond to incidents \cite{shankars2024we,leest2025monitoring}. The challenge of monitoring production ML can therefore be understood as answering two recurring questions (Figure~\ref{fig:overview-monitoring}).

\textit{Q1: How well is the model performing right now, and has performance declined?} \textbf{Performance monitoring requires labels for recent predictions, but in many domains this feedback is delayed or incomplete}. Fraud may be confirmed only after a manual investigation, customer report, or later chargeback \cite{dal2017credit}; churn only after a period of inactivity \cite{runge2014churn}; and loan default only months after a lending decision \cite{li2017reject}. Practitioners therefore rely on feature drift and data integrity metrics as proxies for performance degradation, despite their weak correlation with actual performance loss \cite{leest2025monitoring,shankars2024we}. They need reliable performance estimates to decide whether to intervene -- e.g., by switching to a fallback and warning downstream consumers of the model's predictions.

\textit{Q2: How do the flagged issues affect model performance, and what is worth repairing?} \textbf{Responding to an observed shift requires understanding both its source and its effect on model performance}. Feature drift alone does not identify the source. The population may have changed (e.g., a different mix of customers), or a data integrity issue may have changed how that population is recorded (e.g., a failed table join introducing missing values) \cite{leest2025monitoring,shankars2024we}. These sources can produce the same observed shift but require different responses: a population change may require retraining or recalibration, while a data integrity issue may require repairing the affected table or pipeline \cite{leest2025monitoring}. Data integrity metrics, instrumentation, and lineage can help identify specific issues and the affected artefacts \cite{namakimh2020vamsa, shankars2022towards, heynhm2023automotive}. Even when the source is identified, the magnitude of an observed shift or data integrity violation does not determine how much it affects the model \cite{leest2025monitoring, shankars2024we} -- a join failure making an entire group of columns missing might have little effect, while a short freshness lag in another feature may substantially degrade performance. Engineers thus need to prioritize repairs by their potential recovery of model performance \cite{shankars2024we}.

Both questions require testing how distribution shifts observed in production affect model performance without relying on current label feedback. We study this problem as \emph{online model testing under distribution shift}, referred to simply as \emph{online model testing} throughout this paper. We use these tests to estimate the current model performance (Q1) and the performance gain from repairing individual production issues (Q2). Existing methods typically reproduce distribution shifts under assumptions about their source, such as changes in the population or data integrity issues \cite{decker2024explanatory,chen2021mandoline}. As these sources can occur together in production, representing only one can lead to inaccurate performance estimates when the observed shift also reflects the other (Section~\ref{sec:problem}).

To address this, we examine what must be observed about changes in the population and data pipeline and how these changes can be reproduced in controlled tests. We illustrate the problem and our approach throughout with a churn prediction case for a music streaming service, based on a public dataset~\cite{kkbox}, in which a model predicts whether a user will stop renewing their subscription. Figure~\ref{fig:overview} uses this case to provide an overview of the monitoring setting (Figure~\ref{fig:overview-monitoring}), our framing of the problem of online model testing (Figure~\ref{fig:overview-context}), and the proposed methodology for constructing these tests (Figure~\ref{fig:overview-testing}). Specifically, this paper makes three contributions.

\textbf{We formulate the problem of online model testing under distribution shift (Section~\ref{sec:problem}).} We distinguish changes in the underlying data-generating process from changes in how the data pipeline materializes the observed features and represent them as separate mechanisms (Figure~\ref{fig:overview-context}). This formulation makes explicit that an observed feature shift does not by itself identify which mechanism changed, and that testing its performance effect therefore depends on what can be observed about the underlying mechanism changes and how these changes can be tested.

\textbf{We instantiate this formulation through a methodology for constructing online model tests (Section~\ref{sec:methodology}).} We show how observability of materialization mechanisms can be used to reproduce data integrity issues on labeled reference data and how the effect of changes in the population can be tested despite these issues. We instantiate the methodology for data integrity issues across missingness, freshness, and consistency, specifying what must be observed and how the corresponding changes can be reproduced. Combining these changes provides tests for current model performance (Q1) and the potential gain from repairing a data integrity issue (Q2).

\textbf{We empirically test the proposed formulation through this methodology under controlled mechanism changes (Sections~\ref{sec:experimental_setup}-\ref{sec:case}).} We introduce known materialization changes into real-world datasets with population differences, providing ground truth for their occurrence and performance effects. Across datasets, models, and failure modes, we test whether representing population and materialization changes together improves current performance estimation and whether representing the current population improves repair gain estimation. We additionally assess sensitivity to partial mechanism observability and illustrate the resulting tests in a churn prediction case.

\input{sections/figures/overview_v2}

The rest of the paper is organized as follows. Section~\ref{sec:related_work} positions these contributions with respect to prior work. Following the contribution sections, Section~\ref{sec:discussion} discusses the findings, implications and directions for future work, Section~\ref{sec:limitations} the threats to validity, and Section~\ref{sec:conclusion} concludes.

\section{Background and Related Work}
\label{sec:related_work}

Machine learning operations (MLOps) is the discipline surrounding the development, deployment, and maintenance of ML systems through automated pipelines, versioning, and continuous monitoring \cite{kreuzberger2023machine,amershi2019software}. Monitoring is central to this workflow, detecting drift in model errors and input features and raising alerts that prompt investigation, pipeline repairs, or model retraining \cite{breck2017ml,shankars2024we,leest2025monitoring,zhoux2019a,naveed2025monitoring}. 

\subsection{Distribution Shifts and Data Quality Issues in Production ML}
Data mining research has extensively studied detecting distribution shifts in data streams to monitor ML models \cite{gama2014survey,lu2018,webb2016characterizing,vzliobaite2015overview,bifet2007learning}. For observed features $X$ and outcomes $Y$, \emph{feature drift} is a change in $P(X)$ and \emph{concept drift} a change in $P(Y \mid X)$, detected by two-sample tests on features or prediction errors, respectively \cite{gama2014survey,bifet2007learning}.

Data quality monitoring in ML pipelines builds on frameworks that distinguish dimensions of data quality, e.g., completeness, freshness, and consistency \cite{batini2009methodologies,schelter2018automating,breck2019data,cavenesse2020tensorflow}. Data validation has been studied extensively in database research, where consistency is often checked through constraints, commonly as functional dependencies stating that one column determines another \cite{bohannon2007conditional,chu2013discovering,papenbrock2015functional,rekatsinas2017holoclean}.

\subsection{Testing ML Systems}
Testing ML systems is well established, but is performed predominantly before deployment \cite{zhang2020machine,riccio2020testing}, where expectations about correct behavior are derived from input transformations rather than from labels \cite{xie2011testing,ribeiro2020beyond}. Related studies test model performance under faults or shifts, and methods for their detection and repair. This includes injecting data errors into the data a model is served \cite{schelters2021jenga,schelters2020learning, jager2024data,li2021cleanml}, generating and evaluating alternative pipeline versions \cite{grafberger2023automating}, or simulating changes in the mixture of the data \cite{galewis2022augur, sobolewskip2017scr}. In these studies, the failure modes and shift sources are enumerated at design time, while we test those as they occur and are observed at runtime.

After deployment, online testing is often used as assessment on live traffic \cite{zhang2020machine}, through controlled experiments such as A/B tests and canary releases \cite{kohavi2020trustworthy,auer2021controlled}. 
We extend this notion of online testing to the distribution shifts observed in production to determine their effect on model performance.

\subsection{Performance Estimation Without Labeled Feedback}
When current labels are absent, the performance of a deployed model on unlabeled evaluation data $E$ can be estimated using labeled reference data $R$. Importance weighting (IW) reweighs the labeled observations using the ratio of evaluation to reference feature densities, giving more weight to feature values more prevalent in the current batch \cite{sugiyama2007covariate,chen2021mandoline}. Performance is then computed on the weighted reference observations, which approximate the current feature distribution. This follows the common assumption of performance estimation methods that the shift is in the feature distribution -- $P(Y\mid X)$ remains unchanged. Weights can also be estimated from slices that capture shifts relevant to model loss, avoiding estimation over all features. Other methods estimate accuracy from confidence scores fitted on labeled reference data~\cite{garg2022leveraging}, estimate classification performance from predictions and probability estimates~\cite{bialek2026estimating}, or train a classifier directly to predict performance by enumerating and simulating shifts at design time~\cite{schelters2020learning}. Optimal transport (OT) can be used to estimate performance changes under ``label-preserving'' shifts -- analogous to data integrity changes, explained in Section~\ref{sec:problem} ~\cite{decker2024explanatory}. Performance estimation has thus far been primarily studied under strong assumptions of where the change originates. We address settings in which population change and data quality issues occur together.

\subsection{Diagnosing Faults in ML Systems and Assessing Their Effects}

Root-cause analysis in ML pipelines relies on lineage, metadata recording the sources and transformations through which each feature is produced \cite{namakimh2020vamsa}. Combined with pipeline instrumentation and execution metadata, issues can be traced to the upstream table, transformation, or pipeline run that introduced it \cite{shankars2022towards,leest2025monitoring}. This locates the component responsible, but not the effect on the model.

Causal methods attribute distribution shifts to changes in data-generating mechanisms~\cite{budhathokik2021why}, with IW and Shapley values used to attribute model performance changes given a known causal graph, and labeled data from both environments~\cite{zhangh2023why}. Similar attribution has been applied over software dataflow graphs (in a non-ML application) where computations are treated as mechanisms~\cite{paleyes2023dataflow}. A conceptual notation connects external data-generating processes with internal processing through causal maps of ML systems~\cite{leest2026explaining}. We consider the data-generating process as a single mechanism and distinguish it from changes in individual pipeline mechanisms that materialize features. Other methods operate without causal assumptions, decomposing performance changes into feature and concept drift contributions~\cite{cai2026diagnosing} or attributing estimated changes to individual input features under ``label-preserving'' shifts~\cite{decker2024explanatory}. These also do not explicitly distinguish population and materialization changes, although both can affect the same observed features. We instead test both jointly to assess how data integrity issues affect model performance under a changing population.

\section{Problem}
\label{sec:problem}

We consider a model $f$ trained on historical data and deployed for inference. At each invocation, the recorded feature vector $X^\ast$ and resulting prediction $\hat{Y}=f(X^\ast)$ form an observation in the production data stream. The corresponding label $Y^\ast$ becomes available only after the outcome can be determined and may therefore be delayed or remain absent.

A reference batch \(R\) -- held-out test data or data from an earlier period with acceptable performance -- provides a baseline where labels are available and model performance is known. An evaluation batch \(E\) contains observations from deployment, e.g., data from a later period, for which labels are typically not yet available. A distribution shift occurs when the data in \(E\) follow a different distribution than the data in \(R\). In a monitoring setup, the data stream is partitioned into consecutive windows, such as days or weeks, each serving as a separate evaluation batch.

\subsection{Sources of Distribution Shifts: Data Generating vs Data Materialization Processes}

Distribution shifts can originate in the modeled population or in the data pipeline that produces model features. To separate the two, we borrow the \textit{measurement assumption} from statistics, i.e. the distinction between a true value as it occurs in the population and the measured value \cite{saeed2020anchored,kuroki2014measurement}, and extend it to the data materialization process within an ML system. Figure~\ref{fig:system-context} shows the mechanisms that can change independently and cause distribution shifts. The \emph{data-generating process} $P(X,Y)$ produces the actual features $X$ and outcomes $Y$ as they occur in the population. These are the result of two mechanisms: the marginal distribution $P(X)$ and the conditional distribution $P(Y \mid X)$.
The \emph{data-materialization process} then produces the observed data through feature materialization $P(X^\ast \mid X)$ and label materialization $P(Y^\ast \mid Y)$ -- which produce the features $X^\ast$ and labels $Y^\ast$.

\definecolor{headgreen}{RGB}{228,255,219}
\definecolor{headblue}{RGB}{205,243,255}
\definecolor{nodegreen}{RGB}{217,242,208}
\definecolor{nodeblue}{RGB}{205,243,255}
\definecolor{panelgray}{RGB}{246,246,246}
\definecolor{linecharcoal}{RGB}{48,48,48}
\definecolor{linemid}{RGB}{72,72,72}
\definecolor{glosstext}{RGB}{70,70,70}

\newcommand{\ScenarioTextFont}{\fontsize{8.0}{10.4}\selectfont}
\newcommand{\PanelYUnit}{1.140}
\newcommand{\SlotMechWidth}{18.6mm}
\newcommand{\SlotMechHeight}{7.00mm}
\newcommand{\CompMechWidth}{14.8mm}
\newcommand{\CompMechHeight}{5.20mm}
\newcommand{\MechName}[1]{\boldsymbol{#1}}
\providecommand{\mech}[1]{\ensuremath{\boldsymbol{#1}}}
\newcommand{\MechSymbolFont}{\fontsize{10.60}{10.60}\selectfont}
\newcommand{\MechGlossFont}{\fontsize{8.40}{8.40}\selectfont}
\newcommand{\NamedMech}[2]{%
  {\MechSymbolFont$\MechName{#1}$}{\MechGlossFont\textcolor{glosstext}{: $#2$}}}
\newcommand{\GlossMech}[1]{{\MechGlossFont\textcolor{glosstext}{$#1$}}}
\newcommand{\DomainVariableFont}{\fontsize{9.20}{10.45}\selectfont}
\newcommand{\PanelHeaderFont}{\fontsize{8.35}{9.55}\selectfont}
\newcommand{\RowLabelFont}{\fontsize{8.30}{9.50}\selectfont\itshape}
\newcommand{\RowExampleFont}{\fontsize{6.75}{7.85}\selectfont\itshape}
\newcommand{\SourceBoxWidth}{7.10mm}
\newcommand{\SourceBoxHeight}{6.00mm}
\newcommand{\SourceFormulaFont}{\MechSymbolFont}
\newcommand{\ScenarioHeaderFont}{\fontsize{8.55}{9.75}\selectfont}
\newcommand{\InequalityFont}{\fontsize{9.20}{9.20}\selectfont}
\newcommand{\FlowLineWidth}{.42pt}
\newcommand{\FlowHeadLength}{1.56mm}
\newcommand{\FlowHeadWidth}{1.02mm}
\newcommand{\InnerGenerationLineWidth}{.24pt}
\newcommand{\FeatureMechanismLineWidth}{.37pt}
\newcommand{\SecondaryMechanismLineWidth}{.24pt}
\newcommand{\SourceMechanismLineWidth}{.33pt}
\newcommand{\JointMechanismLineWidth}{.41pt}
\newcommand{\OuterRuleWidth}{.62pt}
\newcommand{\OuterRuleColor}{linemid}
\newcommand{\StructuralRuleWidth}{.36pt}
\newcommand{\DashedSeparatorWidth}{.31pt}
\newcommand{\JointMechanismLeft}{1.96}
\newcommand{\JointMechanismRight}{3.82}
\newcommand{\JointMechanismBottom}{.33}
\newcommand{\JointMechanismTop}{2.42}
\newcommand{\ColumnMechX}{2.89}
\newcommand{\DomainVarX}{4.487}
\newcommand{\GLabelX}{2.89}
\newcommand{\GLabelY}{1.50}
\newcommand{\GLabelAnchor}{center}
\newcommand{\FeedbackTurnY}{1.24}
\newcommand{\FeedbackDropX}{2.89}
\newcommand{\FeedbackFlow}{%
  \draw[flow] (\DomainVarX,1.80)--(\DomainVarX,\FeedbackTurnY)
    --(\FeedbackDropX,\FeedbackTurnY)
    --(\FeedbackDropX,\FeedbackTurnY|-pyx.north);}
\newcommand{\ScenarioCellRight}{4.21}
\newcommand{\SourceCellLeft}{4.45}
\newcommand{\SourceCellRight}{6.54}
\newcommand{\ScenarioPanelWidth}{6.61}
\newcommand{\ScenarioHeaderX}{2.135}
\newcommand{\SourceColumnCenter}{5.495}
\newcommand{\SourceReferenceX}{4.917}
\newcommand{\SourceEvaluationX}{6.073}
\newcommand{\ScenarioTextWidth}{37.0mm}
\newcommand{\NaturalScenarioText}{\frenchspacing%
  Over time, \mbox{\textit{User Activity}} becomes less frequent and
  \textit{churn} increases as \textbf{behavior changes}.}
\newcommand{\JoinScenarioText}{\frenchspacing%
  A \textbf{join failure} omits events from aggregation;
  some \mbox{\textit{User Activity}} counts are recorded as zero.}

\tikzset{
  flow/.style={
    draw=linecharcoal,
    line width=\FlowLineWidth,
    line cap=round,
    line join=round,
    -{Latex[length=\FlowHeadLength,width=\FlowHeadWidth]}
  },
  innergeneration/.style={
    draw=linemid,
    fill=nodegreen,
    line width=\InnerGenerationLineWidth,
    dash pattern=on 1.35pt off 1.05pt,
    rounded corners=1.55pt,
    minimum width=\CompMechWidth,
    minimum height=\CompMechHeight,
    inner xsep=.45pt,
    inner ysep=.55pt
  },
  featurematerialization/.style={
    draw=linecharcoal,
    fill=nodeblue,
    line width=\FeatureMechanismLineWidth,
    rounded corners=1.75pt,
    minimum width=\SlotMechWidth,
    minimum height=\SlotMechHeight,
    inner xsep=.60pt,
    inner ysep=.55pt
  },
  labelmaterialization/.style={
    featurematerialization,
    draw=linemid,
    fill=none,
    line width=\SecondaryMechanismLineWidth,
    dash pattern=on 1.35pt off 1.05pt,
    rounded corners=1.55pt,
    minimum width=\CompMechWidth,
    minimum height=\CompMechHeight
  },
  domainvariable/.style={
    inner sep=0pt,
    font=\DomainVariableFont
  },
  sourcebox/.style={
    draw=linecharcoal,
    line width=\SourceMechanismLineWidth,
    rounded corners=1.55pt,
    minimum width=\SourceBoxWidth,
    minimum height=\SourceBoxHeight,
    inner xsep=.20pt,
    inner ysep=.25pt,
    font=\SourceFormulaFont
  },
  mechanismlabel/.style={
    inner sep=0pt
  }
}

\newcommand{\SystemContextPanel}{%
\begin{tikzpicture}[x=1cm,y=\PanelYUnit cm]
  \path[use as bounding box] (0,0) rectangle (8.23,3.04);

  \fill[headgreen] (1.72,2.69) rectangle (4.975,3.04);
  \fill[headblue]  (4.975,2.69) rectangle (8.23,3.04);
  \draw[draw=\OuterRuleColor,line width=\OuterRuleWidth] (0,3.04)--(8.23,3.04);
  \draw[draw=linemid,line width=\StructuralRuleWidth] (0,2.69)--(8.23,2.69);
  \draw[draw=linemid,line width=\StructuralRuleWidth] (4.975,2.69)--(4.975,3.04);
  \draw[draw=linemid,line width=\StructuralRuleWidth] (1.72,.06)--(1.72,2.69);
  \draw[draw=linemid,line width=\StructuralRuleWidth] (0,1.375)--(1.72,1.375);
  \draw[draw=\OuterRuleColor,line width=\OuterRuleWidth] (0,.06)--(8.23,.06);

  \node[font=\PanelHeaderFont]
    at (3.348,2.865) {Data-Generation};
  \node[font=\PanelHeaderFont]
    at (6.603,2.865) {Data-Materialization};

  \node[font=\RowLabelFont]
    at (.86,2.28) {$X$: Features};
  \draw[draw=linemid,line width=\DashedSeparatorWidth,dash pattern=on 2.0pt off 1.6pt]
    (0,1.96)--(1.72,1.96);
  \node[font=\RowExampleFont]
    at (.86,1.71) {e.g. User Activity};

  \node[font=\RowLabelFont]
    at (.86,.965) {$Y$: Outcome};
  \draw[draw=linemid,line width=\DashedSeparatorWidth,dash pattern=on 2.0pt off 1.6pt]
    (0,.645)--(1.72,.645);
  \node[font=\RowExampleFont]
    at (.86,.395) {e.g. Churn};

  \path[fill=panelgray,rounded corners=1.7pt]
    (1.82,1.49) rectangle (4.875,2.59);
  \path[fill=panelgray,rounded corners=1.7pt]
    (5.075,1.49) rectangle (8.13,2.59);
  \path[fill=panelgray,rounded corners=1.7pt]
    (1.82,.16) rectangle (4.875,1.26);
  \path[fill=panelgray,rounded corners=1.7pt]
    (5.075,.16) rectangle (8.13,1.26);

  \path[
    draw=linecharcoal,fill=nodegreen,line width=\JointMechanismLineWidth,
    rounded corners=2.1pt
  ] (\JointMechanismLeft,\JointMechanismBottom)
    rectangle (\JointMechanismRight,\JointMechanismTop);
  \node[innergeneration] (px) at (\ColumnMechX,2.04) {};
  \node[mechanismlabel,yshift=-.55pt] at (\ColumnMechX,2.04) {\GlossMech{P(X)}};
  \node[mechanismlabel,anchor=\GLabelAnchor,yshift=-.55pt]
    at (\GLabelX,\GLabelY) {\NamedMech{G}{P(X,Y)}};
  \node[innergeneration] (pyx) at (\ColumnMechX,.71) {};
  \node[mechanismlabel,yshift=-.55pt] at (\ColumnMechX,.71) {\GlossMech{P(Y|X)}};

  \node[domainvariable] (x) at (\DomainVarX,2.04) {$X$};
  \node[domainvariable] (y) at (\DomainVarX,.71) {$Y$};
  \node[featurematerialization] (mx) at (6.145,2.04) {};
  \node[mechanismlabel,yshift=-.55pt]
    at (6.145,2.04) {\NamedMech{M}{P(X^\ast|X)}};
  \node[labelmaterialization] (my) at (6.145,.71) {};
  \node[mechanismlabel,yshift=-.55pt]
    at (6.145,.71) {\GlossMech{P(Y^\ast|Y)}};
  \node[domainvariable] (xb) at (7.806,2.04) {$X^\ast$};
  \node[domainvariable] (yb) at (7.806,.71) {$Y^\ast$};

  \draw[flow] (px)--(x);
  \draw[flow] (x)--(mx);
  \draw[flow] (mx)--(xb);
  \draw[flow] (pyx)--(y);
  \draw[flow] (y)--(my);
  \draw[flow] (my)--(yb);
  \FeedbackFlow
\end{tikzpicture}%
}

\newcommand{\ScenarioTable}{%
\begin{tikzpicture}[x=1cm,y=\PanelYUnit cm]
  \path[use as bounding box] (0,0) rectangle (\ScenarioPanelWidth,3.04);

  \path[fill=panelgray,rounded corners=1.7pt]
    (.06,1.49) rectangle (\ScenarioCellRight,2.59);
  \path[fill=panelgray,rounded corners=1.7pt]
    (\SourceCellLeft,1.49) rectangle (\SourceCellRight,2.59);
  \path[fill=panelgray,rounded corners=1.7pt]
    (.06,.16) rectangle (\ScenarioCellRight,1.26);
  \path[fill=panelgray,rounded corners=1.7pt]
    (\SourceCellLeft,.16) rectangle (\SourceCellRight,1.26);

  \draw[draw=\OuterRuleColor,line width=\OuterRuleWidth] (0,3.04)--(\ScenarioPanelWidth,3.04);
  \draw[draw=linemid,line width=\StructuralRuleWidth] (0,2.69)--(\ScenarioPanelWidth,2.69);
  \draw[draw=linemid,line width=\StructuralRuleWidth] (0,1.375)--(\ScenarioPanelWidth,1.375);
  \draw[draw=\OuterRuleColor,line width=\OuterRuleWidth] (0,.06)--(\ScenarioPanelWidth,.06);

  \node[font=\ScenarioHeaderFont]
    at (\ScenarioHeaderX,2.865) {Shift Scenario};
  \node[font=\ScenarioHeaderFont]
    at (\SourceColumnCenter,2.865) {Shift Source};

  \node[
    anchor=west,
    align=justify,
    inner sep=0pt,
    text width=\ScenarioTextWidth,
    font=\ScenarioTextFont
  ] at (.285,2.04)
    {\NaturalScenarioText};
  \node[sourcebox,fill=nodegreen] at (\SourceReferenceX,2.04)
    {$\MechName{G}^{R}$};
  \node[inner sep=1.0pt,
    font=\InequalityFont] at (\SourceColumnCenter,2.04)
    {$\ne$};
  \node[sourcebox,fill=nodegreen] at (\SourceEvaluationX,2.04)
    {$\MechName{G}^{E}$};

  \node[
    anchor=west,
    align=justify,
    inner sep=0pt,
    text width=\ScenarioTextWidth,
    font=\ScenarioTextFont
  ] at (.285,.71)
    {\JoinScenarioText};
  \node[sourcebox,fill=nodeblue] at (\SourceReferenceX,.71)
    {$\MechName{M}^{R}$};
  \node[inner sep=1.0pt,
    font=\InequalityFont] at (\SourceColumnCenter,.71)
    {$\ne$};
  \node[sourcebox,fill=nodeblue] at (\SourceEvaluationX,.71)
    {$\MechName{M}^{E}$};
\end{tikzpicture}%
}

\newcommand{\SystemContextCapA}{The data-generating process \mech{G} produces the actual $X$
and $Y$, which materialization \mech{M} produces as $X^\ast$ and
$Y^\ast$.}
\newcommand{\SystemContextCapB}{Both scenarios shift $X^\ast$, but one changes
\mech{G} and the other \mech{M} between batches $R$ and $E$.}
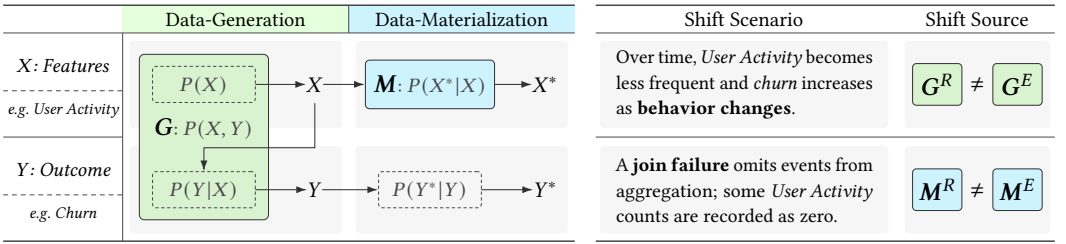
\newcommand{\SystemContextCapMain}{Sources of distribution shifts; the mechanisms in which they originate (a) and example scenarios (b).}

\ifdefined\SystemContextDefinitionsOnly
\else

\begin{figure*}[b]
  \centering
  \begin{subfigure}[b]{.5435\textwidth}
    \vspace{0pt}
    \centering
    \resizebox{\linewidth}{!}{\SystemContextPanel}
    \caption{\SystemContextCapA}
    \label{fig:system-context}
  \end{subfigure}%
  \hfill
  \begin{subfigure}[b]{.4365\textwidth}
    \vspace{0pt}
    \centering
    \resizebox{\linewidth}{!}{\ScenarioTable}
    \caption{\SystemContextCapB}
    \label{fig:shift-scenarios}
  \end{subfigure}
  \caption{\SystemContextCapMain}
  \label{fig:system-context-and-shift-scenarios}
\end{figure*}

\fi

We focus on the data-generating process and the feature materialization mechanism, because both can degrade model performance when they change. Label materialization affects only how we measure that performance and falls outside our scope.

Figure~\ref{fig:shift-scenarios} shows how the different sources can have the same effect on the churn prediction example introduced in Section~\ref{sec:introduction}: \textit{Suppose user activity features are computed by aggregating user events. When users interact with the service less often, their observed activity declines. Because lower activity in the population is associated with a higher probability of churn, the actual churn rate increases. A failed join can produce the same decline in observed activity when an identifier mismatch prevents the aggregation from matching some users to their events, causing their activity counts to be filled with zero. However, their actual activity and thus churn risk remain unchanged. The first shift therefore originates in the population $P(X,Y)$ and raises the marginal probability of churn $P(Y)$, while the second originates in feature materialization $P(X^\ast\mid X)$ and leaves it unchanged}.

\subsection{Limitations of Online Model Testing When Shift Sources Are Assumed}

The source of an observed shift determines how its effect on model performance can be tested. Importance weighting (IW) \cite{sugiyama2007covariate,chen2021mandoline} and optimal transport (OT) \cite{decker2024explanatory} for performance estimation (Section~\ref{sec:related_work}) can be understood as assuming a particular source of the shift in $E$ and reproducing the corresponding mechanism change on the labeled reference data $R$. Since the observed distribution alone does not identify which mechanism changed, reproducing the shift as a change in the wrong mechanism can bias the estimates. For illustration, we estimate the F1-score of a LightGBM classifier predicting churn on $E$ under three scenarios. A \emph{behavior change} (in the population), simulated by upsampling users with lower activity; an emulated \emph{join failure} (materialization change), in which activity features are zeroed for a random 50\% of rows; and a \emph{mixed shift} combining both. Figure~\ref{fig:conflation-showcase} shows the absolute errors of IW and OT against a baseline that assumes no change in F1 score.

\noindent\textbf{Materialization change assumed to be population change}. IW estimates performance in $E$ by reweighting labeled $R$ so its observed feature distribution $P(X^\ast)$ matches that of $E$, relying on the covariate shift assumption that $P(Y \mid X^\ast)$ is invariant between $R$ and $E$. Under \textit{behavior change} this invariance might very well hold. In the \textit{join failure} scenario, however, many users assigned zero activity in $E$ remain active and have lower churn risk than genuinely inactive users with zero activity in $R$. This causes a change in $P(Y\mid X^\ast)$, appearing as concept drift even though $P(Y\mid X)$ remains stable, and reweighting the zero-activity users in $R$ therefore overweights users with higher churn risk. Accordingly, Figure~\ref{fig:conflation-showcase} shows that IW estimates performance accurately under population change, while its error is substantially larger under \textit{join failure} and \textit{mixed shift} compared to the baseline.

\providecommand{\MechName}[1]{\boldsymbol{#1}}
\providecommand{\mech}[1]{\ensuremath{\boldsymbol{#1}}}
\providecommand{\OuterRuleWidth}{.62pt}
\providecommand{\StructuralRuleWidth}{.36pt}
\providecommand{\SourceMechanismLineWidth}{.33pt}
\providecommand{\SecondaryMechanismLineWidth}{.24pt}

\definecolor{nodegreen}{RGB}{217,242,208}
\definecolor{nodeblue}{RGB}{205,243,255}
\definecolor{linecharcoal}{RGB}{48,48,48}
\definecolor{linemid}{RGB}{72,72,72}

\definecolor{softdiv}{RGB}{198,198,198}   %
\definecolor{barBody}{RGB}{207,207,207}   %
\definecolor{barOver}{RGB}{251,238,236}   %
\definecolor{barStub}{RGB}{247,247,247}   %
\definecolor{tickInk}{RGB}{58,58,58}
\definecolor{whiskInk}{RGB}{51,51,51}
\definecolor{numInk}{RGB}{26,26,26}
\definecolor{redNum}{RGB}{179,38,30}      %

\newcommand{\DivColor}{softdiv}           %
\newcommand{\DivWidth}{.4pt}              %

\newlength{\ShowTrackW}\setlength{\ShowTrackW}{23mm}  %
\newlength{\ShowGapW}  \setlength{\ShowGapW}{1.6mm}   %
\newlength{\ShowNumW}  \setlength{\ShowNumW}{8mm}     %
\newlength{\ShowCellW}
\setlength{\ShowCellW}{\ShowTrackW}
\addtolength{\ShowCellW}{\ShowGapW}
\addtolength{\ShowCellW}{\ShowNumW}
\pgfmathsetmacro{\ShowCellX}{\ShowCellW/\ShowTrackW}  %
\pgfmathsetmacro{\ShowCellMid}{\ShowCellX/2}          %
\newcommand{\BarH}{1.00}          %
\newcommand{\CapH}{0.95}          %
\newcommand{\TickH}{1.85}         %
\newcommand{\HeadFont}{\footnotesize}

\tikzset{
  showchip/.style    ={rounded corners=1.55pt,inner xsep=2.4pt,inner ysep=1.4pt,
                       minimum width=4.6mm},
  showchip on/.style ={showchip,draw=linecharcoal,
                       line width=\SourceMechanismLineWidth},
  showchip off/.style={showchip,draw=linemid,fill=none,text=black!45,
                       line width=\SecondaryMechanismLineWidth,
                       dash pattern=on 1.35pt off 1.05pt},
  showcell/.style    ={x=\ShowTrackW,y=1mm,baseline=-0.62ex,
                       line cap=butt,line join=miter},
  showhead/.style    ={x=\ShowTrackW,y=1mm,baseline=0pt},
}
\newcommand{\ShowChipG}[1]{\ifnum#1=1
  \tikz[baseline=(c.base)]{\node[showchip on,fill=nodegreen](c){$\MechName{G}$};}\else
  \tikz[baseline=(c.base)]{\node[showchip off](c){$\MechName{G}$};}\fi}
\newcommand{\ShowChipM}[1]{\ifnum#1=1
  \tikz[baseline=(c.base)]{\node[showchip on,fill=nodeblue](c){$\MechName{M}$};}\else
  \tikz[baseline=(c.base)]{\node[showchip off](c){$\MechName{M}$};}\fi}
\newcommand{\ShowMech}[2]{\ShowChipG{#1}\,\ShowChipM{#2}}

\newcommand{\ShowRule}[1]{%
  \noalign{\vskip3pt{\color{linemid}\hrule height #1}\vskip3pt}}
\newcommand{\ShowDiv}{%
  \noalign{\vskip2.2pt{\color{\DivColor}\hrule height \DivWidth}\vskip2.2pt}}

\newcommand{\ShowHead}[1]{%
  \begin{tikzpicture}[showhead]
    \useasboundingbox (0,-0.60) rectangle (\ShowCellX,2.10);
    \node[anchor=base,inner sep=0pt,font=\HeadFont] at (\ShowCellMid,0) {#1};
  \end{tikzpicture}}

\newcommand{\ShowCell}[6]{%
  \begin{tikzpicture}[showcell]
    \useasboundingbox (0,-2.15) rectangle (\ShowCellX,2.15);
    \pgfmathsetmacro{\um}{#1/#4}\pgfmathsetmacro{\ug}{#5/#4}
    \pgfmathsetmacro{\ua}{#2/#4}\pgfmathsetmacro{\ub}{#3/#4}
    \pgfmathparse{#1>#5 ? 1 : 0}\edef\isbad{\pgfmathresult}
    \ifnum\isbad=1
      \path[fill=barBody] (0,-\BarH) rectangle (\ug,\BarH);
      \path[fill=barOver] (\ug,-\BarH) rectangle (\um,\BarH);
      \def\nc{redNum}
    \else
      \path[fill=barBody] (0,-\BarH) rectangle (\um,\BarH);
      \path[fill=barStub] (\um,-\BarH) rectangle (\ug,\BarH);
      \def\nc{numInk}
    \fi
    \draw[tickInk,line width=.7pt] (\ug,-\TickH) -- (\ug,\TickH);
    \draw[whiskInk,line width=.5pt,line cap=round] (\ua,0) -- (\ub,0);
    \draw[whiskInk,line width=.5pt] (\ua,-\CapH) -- (\ua,\CapH);
    \draw[whiskInk,line width=.5pt] (\ub,-\CapH) -- (\ub,\CapH);
    \node[anchor=east,inner sep=0pt,text=\nc] at (\ShowCellX,0) {#6};
  \end{tikzpicture}}

\newcommand{\ShowCols}{%
  & \ShowHead{No change (baseline)}
  & \ShowHead{Importance weighting (IW)}
  & \ShowHead{Optimal transport (OT)} \\}

\newcommand{\ConflationPanel}{%
  \small
  \setlength{\tabcolsep}{0pt}%
  \setlength{\arrayrulewidth}{0.25pt}%
  \begin{tabular*}{\linewidth}
  {@{}l@{\hspace{6pt}}l@{\extracolsep{\fill}}%
      !{\hspace{2.5pt}{\color{\DivColor}\vrule width \DivWidth}\hspace{2.5pt}}c%
      !{\hspace{2.5pt}{\color{\DivColor}\vrule width \DivWidth}\hspace{2.5pt}}c%
      !{\hspace{2.5pt}{\color{\DivColor}\vrule width \DivWidth}\hspace{2.5pt}}c@{}}
    \cline{1-5}\noalign{\vskip3pt}%
    & & \multicolumn{3}{c}{Estimation error $\downarrow$ -- estimated vs. actual F1-score} \\[1.5pt]
    \cline{3-5}\noalign{\vskip3pt}%
    & Shift scenario \ShowCols
    \ShowDiv
    \ShowMech{1}{0} & Behavior Change
      & \ShowCell{0.123063343716}{0.092309853808}{0.161591512400}{0.445037275128}{0.123063343716}{0.12}
      & \ShowCell{0.012327487084}{0.002460442612}{0.030166600382}{0.445037275128}{0.123063343716}{0.01}
      & \ShowCell{0.407786389971}{0.364762055310}{0.445037275128}{0.445037275128}{0.123063343716}{0.41} \\
    \ShowDiv
    \ShowMech{0}{1} & Join failure
      & \ShowCell{0.088493708263}{0.064899616811}{0.101329458457}{0.177927016419}{0.088493708263}{0.09}
      & \ShowCell{0.162431424544}{0.137470117739}{0.177927016419}{0.177927016419}{0.088493708263}{0.16}
      & \ShowCell{0.049197457625}{0.032921810700}{0.063786008230}{0.177927016419}{0.088493708263}{0.05} \\
    \ShowDiv
    \ShowMech{1}{1} & Mixed shift
      & \ShowCell{0.055138966282}{0.011642952730}{0.083082927187}{0.399158307391}{0.055138966282}{0.06}
      & \ShowCell{0.107657601642}{0.090205793015}{0.135651743079}{0.399158307391}{0.055138966282}{0.11}
      & \ShowCell{0.364673229128}{0.309936486073}{0.399158307391}{0.399158307391}{0.055138966282}{0.36} \\
    \ShowRule{\OuterRuleWidth}
  \end{tabular*}}

\begin{figure}[b]
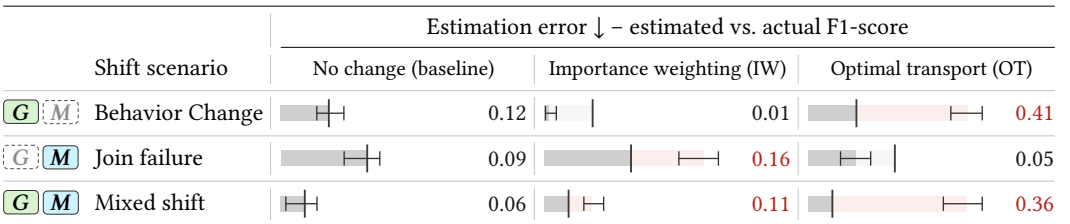

  \centering
  \ConflationPanel
  \vspace{-0.8em}
  \caption{Each estimator is accurate only where the mechanism it transports
    is the one that changed (\mech{G} or \mech{M}); under mixed shift neither
    is. Bars show the mean absolute error (MAE) over 30 draws with min--max
    bounds.}
  \label{fig:conflation-showcase}
\end{figure}

\noindent\textbf{Population change assumed to be materialization change}. OT estimates labels in $E$ by pairing each evaluation feature vector with a labeled reference observation, assuming the shift changes observed features but preserves corresponding outcomes -- as a change in materialization $P(X^\ast\mid X)$ would. The \textit{join failure} is such a shift, changing the activity presented to the model without affecting actual activity or churn risk, and OT reproduces it by moving activity values in $R$ towards the $E$ distribution while retaining the reference labels. Under \textit{behavior change}, the same operation reproduces the shift as a change in the wrong mechanism, because users behind the lower activity in $E$ are more likely to churn. Figure~\ref{fig:conflation-showcase} confirms this; the estimation error of OT falls below the baseline under \textit{join failure}, and rises substantially above it under \textit{behavior change} and \textit{mixed shift}.

\subsection{Observability and Transport for Online Model Testing}

The preceding example shows that an observed feature shift alone does not establish which mechanism change a test should reproduce. Testing therefore requires \textbf{mechanism observability}, determining which selected mechanisms changed and how those changes affect observed data.

The materialization process can be understood as a data flow of source tables, joins, and pipelines whose operations compose the mechanism $P(X^\ast\mid X)$. Instrumenting these operations can expose their individual mechanisms -- a broken pipeline surfaces as a stale aggregation snapshot, faulty enrichment as a join indicator (see Section~\ref{sec:methodology}). Observability of the materialization mechanism is therefore \textit{partial} and \textit{specific} to the instrumented parts of the system. The degree to which we make the materialization mechanisms observable also constrains what we can infer about population change. Since $X$ is observed only through $X^\ast$, inferring population change requires comparable materialization between $R$ and $E$ and is limited to changes represented by the comparable data.

Observing a mechanism change does not yet quantify its performance effect. Testing it requires a \textbf{mechanism transport}, an operation that reproduces the selected change on labeled $R$ while leaving other mechanism states unchanged. Because this effect can depend on the population and other materialization mechanisms (e.g., the same failed join may be harmless in one population but damaging in another), online model testing requires composing the available transports to reproduce the relevant evaluation conditions. This gives the challenge addressed in this paper.

\textbf{How can changes in data-generating and materialization mechanisms be observed and transports constructed and composed to test effects on model performance?}

\section{Proposed Methodology}
\label{sec:methodology}

We instantiate the framing in Section~\ref{sec:problem} by constructing materialization and population transports from observable mechanism changes (Section~\ref{sec:mechanism_transport}) and composing them on labeled reference data to estimate current performance and repair gains (Section~\ref{sec:testing}).

\providecolor{overviewGreenLine}{HTML}{3A9F2D}
\providecolor{overviewBlueLine}{HTML}{62C4E9}
\providecolor{overviewGreen}{HTML}{D9F2D0}
\providecolor{overviewBlue}{HTML}{CAEEFB}
\providecolor{overviewShade}{HTML}{F2F2F2}
\providecolor{mtCell}{HTML}{EDEDED}
\providecolor{mtBlueText}{HTML}{0B5E7E}
\providecolor{mtGreenText}{HTML}{2C7A22}
\tikzset{
  mtstep/.style={inner sep=0pt, font=\fontsize{8.3}{9.5}\selectfont},
  mtchip/.style={draw=black, line width=.45pt, rounded corners=1.2pt, inner xsep=2.2pt,
                 minimum height=.36cm, inner ysep=0pt, font=\fontsize{8}{9}\selectfont}}
\begin{figure*}[b]

  \centering
\begin{tikzpicture}[x=1cm,y=1cm]
  \tikzset{c/.style={fill=mtCell,draw=white,line width=1pt,minimum width=0.60cm,minimum height=0.31cm,inner sep=0pt,font=\fontsize{7.2}{8}\selectfont},
    h/.style={inner sep=0pt,font=\fontsize{7.2}{8}\selectfont\itshape,text=black!75},
    cb/.style={c,fill=overviewBlue,text=mtBlueText},cg/.style={c,fill=overviewGreen,text=mtGreenText},cs/.style={c,fill=white,draw=overviewBlueLine,line width=.5pt,dash pattern=on 1.3pt off 1pt},arr/.style={line width=.55pt,-{Latex[length=1.6mm,width=1.15mm]}},panel/.style={fill=overviewShade,rounded corners=2pt},chipM/.style={mtchip,fill=overviewBlue},chipG/.style={mtchip,fill=overviewGreen},chipOff/.style={mtchip,fill=white,draw=black!30,text=black!40}}
  \path[use as bounding box] (0,0) rectangle (13.90,2.870);
  \tikzset{c/.append style={minimum height=0.370cm,text=black,font=\fontsize{7.2}{8}\selectfont},cg/.append style={text=black},cb/.append style={text=black}}
  \tikzset{chipM/.append style={minimum width=.56cm},chipG/.append style={minimum width=.56cm}}
  \draw[black!80,line width=.62pt] (0,2.850)--(13.90,2.850);
  \node[mtstep,anchor=base] at (2.200,2.570) {\textbf{Evaluation} $E$};
  \node[mtstep,anchor=base] at (6.950,2.570) {\textbf{Transport}};
  \node[mtstep,anchor=base] at (11.700,2.570) {\textbf{Reference} $R$ \textit{after transport}};
  \draw[black!80,line width=.36pt] (0,2.480)--(13.90,2.480);
  \draw[black!80,line width=.36pt] (4.400,2.480)--(4.400,2.850) (9.500,2.480)--(9.500,2.850);
  \draw[black!80,line width=.62pt] (0,.02)--(13.90,.02);
  \node[h] at (1.010,2.175) {$I_M$};
  \node[h] at (1.690,2.175) {\textit{Age}};
  \node[h] at (2.460,2.175) {\textit{Ctry}};
  \node[h] at (3.000,2.175) {\textit{Act.}};
  \node[h] at (3.470,2.175) {$Y$};
  \node[h] at (10.510,2.175) {$I_M$};
  \node[h] at (11.190,2.175) {\textit{Age}};
  \node[h] at (11.960,2.175) {\textit{Ctry}};
  \node[h] at (12.500,2.175) {\textit{Act.}};
  \node[h] at (12.970,2.175) {$Y$};
  \node[h] at (13.580,2.175) {$w$};
  \draw[black!50,line width=.45pt] (10.270,2.345) -- (10.270,2.405) -- (11.550,2.405) -- (11.550,2.345);
  \node[cb,minimum width=0.560cm] at (1.010,1.805) {1};
  \node[cb,minimum width=0.800cm] at (1.690,1.805) {33};
  \node[c,minimum width=0.540cm] at (2.460,1.805) {DE};
  \node[c,minimum width=0.540cm] at (3.000,1.805) {7};
  \node[c,fill=overviewShade,text=black!35,minimum width=0.400cm] at (3.470,1.805) {?};
  \node[cs,minimum width=0.560cm] at (10.510,1.805) {0};
  \node[c,minimum width=0.800cm] at (11.190,1.805) {34};
  \node[c,minimum width=0.540cm] at (11.960,1.805) {NL};
  \node[c,minimum width=0.540cm] at (12.500,1.805) {21};
  \node[c,minimum width=0.400cm] at (12.970,1.805) {0};
  \node[cg,minimum width=0.620cm] at (13.580,1.805) {.55};
  \node[c,fill=white,draw=white,minimum width=0.560cm] at (1.010,1.435) {0};
  \node[c,minimum width=0.800cm] at (1.690,1.435) {41};
  \node[c,minimum width=0.540cm] at (2.460,1.435) {NL};
  \node[c,minimum width=0.540cm] at (3.000,1.435) {3};
  \node[c,fill=overviewShade,text=black!35,minimum width=0.400cm] at (3.470,1.435) {?};
  \node[cs,fill=overviewBlue,minimum width=0.560cm] at (10.510,1.435) {};
  \node[inner sep=0pt,font=\fontsize{7.2}{8}\selectfont] (so) at (10.392,1.435) {0};
  \draw[black!70,line width=.45pt] ([xshift=-.2mm]so.west) -- ([xshift=.2mm]so.east);
  \node[inner sep=0pt,font=\fontsize{7.2}{8}\selectfont] at (10.628,1.435) {1};
  \node[cb,minimum width=0.800cm] at (11.190,1.435) {};
  \node[inner sep=0pt,font=\fontsize{7.2}{8}\selectfont] (so) at (11.014,1.435) {45};
  \draw[black!70,line width=.45pt] (so.west) -- (so.east);
  \node[inner sep=0pt,font=\fontsize{7.2}{8}\selectfont] at (11.382,1.435) {33};
  \node[c,minimum width=0.540cm] at (11.960,1.435) {DE};
  \node[c,minimum width=0.540cm] at (12.500,1.435) {2};
  \node[c,minimum width=0.400cm] at (12.970,1.435) {1};
  \node[cg,minimum width=0.620cm] at (13.580,1.435) {1.35};
  \node[cb,minimum width=0.560cm] at (1.010,1.065) {1};
  \node[cb,minimum width=0.800cm] at (1.690,1.065) {33};
  \node[c,minimum width=0.540cm] at (2.460,1.065) {DE};
  \node[c,minimum width=0.540cm] at (3.000,1.065) {12};
  \node[c,fill=overviewShade,text=black!35,minimum width=0.400cm] at (3.470,1.065) {?};
  \node[cs,fill=overviewBlue,minimum width=0.560cm] at (10.510,1.065) {};
  \node[inner sep=0pt,font=\fontsize{7.2}{8}\selectfont] (so) at (10.392,1.065) {0};
  \draw[black!70,line width=.45pt] ([xshift=-.2mm]so.west) -- ([xshift=.2mm]so.east);
  \node[inner sep=0pt,font=\fontsize{7.2}{8}\selectfont] at (10.628,1.065) {1};
  \node[cb,minimum width=0.800cm] at (11.190,1.065) {};
  \node[inner sep=0pt,font=\fontsize{7.2}{8}\selectfont] (so) at (11.014,1.065) {51};
  \draw[black!70,line width=.45pt] (so.west) -- (so.east);
  \node[inner sep=0pt,font=\fontsize{7.2}{8}\selectfont] at (11.382,1.065) {33};
  \node[c,minimum width=0.540cm] at (11.960,1.065) {NL};
  \node[c,minimum width=0.540cm] at (12.500,1.065) {14};
  \node[c,minimum width=0.400cm] at (12.970,1.065) {0};
  \node[cg,minimum width=0.620cm] at (13.580,1.065) {.70};
  \node[c,fill=white,draw=white,minimum width=0.560cm] at (1.010,0.695) {0};
  \node[c,minimum width=0.800cm] at (1.690,0.695) {24};
  \node[c,minimum width=0.540cm] at (2.460,0.695) {DE};
  \node[c,minimum width=0.540cm] at (3.000,0.695) {4};
  \node[c,fill=overviewShade,text=black!35,minimum width=0.400cm] at (3.470,0.695) {?};
  \node[cs,minimum width=0.560cm] at (10.510,0.695) {0};
  \node[c,minimum width=0.800cm] at (11.190,0.695) {27};
  \node[c,minimum width=0.540cm] at (11.960,0.695) {DE};
  \node[c,minimum width=0.540cm] at (12.500,0.695) {5};
  \node[c,minimum width=0.400cm] at (12.970,0.695) {1};
  \node[cg,minimum width=0.620cm] at (13.580,0.695) {1.65};
  \node[cb,minimum width=0.560cm] at (1.010,0.325) {1};
  \node[cb,minimum width=0.800cm] at (1.690,0.325) {33};
  \node[c,minimum width=0.540cm] at (2.460,0.325) {NL};
  \node[c,minimum width=0.540cm] at (3.000,0.325) {9};
  \node[c,fill=overviewShade,text=black!35,minimum width=0.400cm] at (3.470,0.325) {?};
  \node[cs,fill=overviewBlue,minimum width=0.560cm] at (10.510,0.325) {};
  \node[inner sep=0pt,font=\fontsize{7.2}{8}\selectfont] (so) at (10.392,0.325) {0};
  \draw[black!70,line width=.45pt] ([xshift=-.2mm]so.west) -- ([xshift=.2mm]so.east);
  \node[inner sep=0pt,font=\fontsize{7.2}{8}\selectfont] at (10.628,0.325) {1};
  \node[cb,minimum width=0.800cm] at (11.190,0.325) {};
  \node[inner sep=0pt,font=\fontsize{7.2}{8}\selectfont] (so) at (11.014,0.325) {38};
  \draw[black!70,line width=.45pt] (so.west) -- (so.east);
  \node[inner sep=0pt,font=\fontsize{7.2}{8}\selectfont] at (11.382,0.325) {33};
  \node[c,minimum width=0.540cm] at (11.960,0.325) {NL};
  \node[c,minimum width=0.540cm] at (12.500,0.325) {11};
  \node[c,minimum width=0.400cm] at (12.970,0.325) {0};
  \node[cg,minimum width=0.620cm] at (13.580,0.325) {.75};
  \draw[overviewGreenLine,line width=.6pt,rounded corners=1.5pt] (2.220,0.160) rectangle (3.240,1.970);
  \draw[overviewBlueLine!85!black,line width=.6pt,rounded corners=1.5pt] (0.760,0.160) rectangle (2.060,1.970);
  \path[panel] (4.480,1.005) rectangle (9.420,2.285);
  \node[chipM,anchor=west] at (4.620,2.025) {$M$};
  \node[anchor=west,inner sep=0pt,font=\fontsize{7.8}{9}\selectfont\bfseries] at (5.340,2.025) {Materialization transport};
  \node[anchor=west,inner sep=0pt,font=\fontsize{7.3}{8.4}\selectfont] at (5.340,1.705) {\textit{} \textit{Age} missing in 60\% of $E$};
  \node[anchor=west,inner sep=0pt,font=\fontsize{7.3}{8.4}\selectfont] at (5.340,1.445) {\textit{} Sample $I_M$ at this rate per row};
  \node[anchor=west,inner sep=0pt,font=\fontsize{7.3}{8.4}\selectfont] at (5.340,1.185) {\textit{} Impute \textit{Age} where $I_M{=}1$};
  \path[panel] (4.480,0.085) rectangle (9.420,0.845);
  \node[chipG,anchor=west] at (4.620,0.585) {$G$};
  \node[anchor=west,inner sep=0pt,font=\fontsize{7.8}{9}\selectfont\bfseries] at (5.340,0.585) {Population transport};
  \node[anchor=west,inner sep=0pt,font=\fontsize{7.3}{8.4}\selectfont] at (5.340,0.265) {\textit{reweight} $R$ to match $E$ on \textit{Ctry}, \textit{Act.}};
  \draw[arr,draw=overviewBlueLine!85!black] (3.820,1.645) -- (4.330,1.645);
  \draw[arr,draw=overviewBlueLine!85!black] (9.570,1.645) -- (10.080,1.645);
  \draw[arr,draw=overviewGreenLine] (3.820,0.465) -- (4.330,0.465);
  \draw[arr,draw=overviewGreenLine] (9.570,0.465) -- (10.080,0.465);
\end{tikzpicture}
\vspace{-1.7em}
  \caption{\textbf{Transport.} We reproduce the changes in $E$ on reference data $R$ with churn label $Y$: missing \textit{Age} by sampling $I_M$ at its rate in $E$, and the population change by reweighting on unaffected values (green outline).}
  \label{fig:transport}
\end{figure*}

\subsection{Mechanism Transport}
\label{sec:mechanism_transport}

As established in Section~\ref{sec:problem}, materialization and population transports call for different operations. We explain both below, illustrated on a toy example in Figure~\ref{fig:transport}. A \emph{materialization transport} ($M$) operates on the observed features in $R$, reproducing how selected features are materialized in $E$ while leaving other variables unchanged (Section~\ref{sec:materialization_transport}). A \emph{population transport} ($G$) reweighs observations in $R$ using variables whose materialization remains comparable between $R$ and $E$. Under the assumptions in Section~\ref{sec:population_transport}, this reproduces the part of the population change represented by those variables while preserving their relationship with the remaining features and outcomes.

\subsubsection{Materialization Transport}
\label{sec:materialization_transport}
\label{sec:transport_template}
\label{sec:transport_patterns}

Continuing the churn prediction case, consider a pipeline that retrieves
\textit{user attributes} by joining a user profile table, aggregates
listening events into daily snapshots of \textit{listening activity},
and produces \textit{payment features} by mapping payment method IDs
to payment types and converting currencies. Each of these operations can introduce changes in how the features
are materialized. Profile joins can fail, snapshot refreshes can be
delayed, and the mappings can change. We can make these changes observable through information commonly exposed by modern feature and workflow tooling, such as whether a join succeeded for a particular observation, when a feature was last materialized, and from which source data it was computed~\cite{de2024hopsworks,tecton_http_api,feast_feature_view,dbt_sources}.

\newlength{\obsrulewidth}
\ifdefined\heavyrulewidth\setlength{\obsrulewidth}{\heavyrulewidth}\else\setlength{\obsrulewidth}{.8pt}\fi
\usetikzlibrary{arrows.meta}
\definecolor{headblue}{RGB}{205,243,255}
\definecolor{nodeblue}{RGB}{205,243,255}
\definecolor{panelgray}{RGB}{246,246,246}
\definecolor{linecharcoal}{RGB}{48,48,48}
\definecolor{linemid}{RGB}{72,72,72}
\definecolor{glosstext}{RGB}{70,70,70}
\tikzset{
  obsflow/.style={draw=linecharcoal, line width=.42pt, line cap=round, line join=round,
                  -{Latex[length=1.5mm,width=1.0mm]}},
  obshead/.style={inner sep=0pt, font=\fontsize{8.35}{9.55}\selectfont},
  obsrow/.style={inner sep=0pt, font=\fontsize{8.3}{9.5}\selectfont\itshape}
}

\tikzset{
  ivar/.style={circle, draw=linecharcoal, fill=nodeblue, line width=.37pt,
               minimum size=6.0mm, inner sep=0pt, font=\fontsize{9.0}{10}\selectfont},
  fvar/.style={inner sep=1.6pt, font=\fontsize{9.0}{10}\selectfont},
  ibox/.style={draw=linecharcoal, fill=white, line width=.37pt, rounded corners=1.7pt,
               minimum size=7.6mm, inner sep=0pt},
  nlab/.style={inner sep=0pt, font=\fontsize{6.8}{7.6}\selectfont, text=glosstext},
  elab/.style={inner sep=0pt, font=\fontsize{7.0}{7.8}\selectfont\itshape, text=glosstext},
  clab/.style={inner sep=1pt, font=\fontsize{7.4}{8.2}\selectfont, text=glosstext}
}
\newcommand{\InsNode}[5]{\node[ivar] (#1) at (#2,#3) {#4};\node[nlab,anchor=south] at (#2,#3+.37) {\texttt{#5}};}
\newcommand{\SrcNode}[5]{\node[ivar] (#1) at (#2,#3) {#4};\node[elab,anchor=north] at (#2,#3-.44) {#5};}
\newcommand{\FeatNode}[5]{\node[fvar] (#1) at (#2,#3) {#4};\node[elab,anchor=north] at (#2,#3-.44) {#5};}

\newcommand{\ObsSketch}{%
\begin{tikzpicture}[x=1cm,y=1cm]
  \path[use as bounding box] (0,0) rectangle (13.9,2.99);
  \def\xa{1.72}\def\colw{4.06}\def\off{1.05}
  \def\yI{1.85}\def\yL{.84}
  \draw[draw=linemid,line width=\obsrulewidth] (0,2.99)--(13.9,2.99);
  \draw[draw=linemid,line width=.36pt] (0,2.64)--(13.9,2.64);
  \foreach \k in {1,2}{\draw[draw=linemid,line width=.36pt] (\xa+\k*\colw,2.64)--(\xa+\k*\colw,2.99);}
  \draw[draw=linemid,line width=\obsrulewidth] (0,0)--(13.9,0);
  \node[obshead] at (\xa+.5*\colw,2.815) {Missingness};
  \node[obshead] at (\xa+1.5*\colw,2.815) {Freshness};
  \node[obshead] at (\xa+2.5*\colw,2.815) {Consistency};
  \draw[draw=linemid,line width=.36pt] (\xa,0)--(\xa,2.64);
  \draw[draw=linemid,line width=.36pt] (0,1.35)--(\xa,1.35);
  \node[obsrow,align=center] at (.5*\xa,1.995) {Metadata:\\$I$};
  \node[obsrow,align=center] at (.5*\xa,.675) {Source data:\\$Z$};
  \path[fill=panelgray,rounded corners=1.7pt] (\xa+.08,1.43) rectangle (\xa+\colw-.08,2.57);
  \path[fill=panelgray,rounded corners=1.7pt] (\xa+.5*\colw+.04,.07) rectangle (\xa+\colw-.08,1.27);
  \path[fill=panelgray,rounded corners=1.7pt] (\xa+\colw+.08,1.43) rectangle (\xa+2*\colw-.08,2.57);
  \path[fill=panelgray,rounded corners=1.7pt] (\xa+\colw+.08,.07) rectangle (\xa+2*\colw-.08,1.27);
  \path[fill=panelgray,rounded corners=1.7pt] (\xa+2.5*\colw+.04,1.43) rectangle (\xa+3*\colw-.08,2.57);
  \path[fill=panelgray,rounded corners=1.7pt] (\xa+2*\colw+.08,.07) rectangle (\xa+3*\colw-.08,1.27);
  \begin{scope}[shift={(\xa+.5*\colw,0)}]
    \InsNode{mg}{-\off}{\yI}{$I_J$}{join\_mismatch}
    \InsNode{m}{\off}{\yI}{$I_M$}{is\_missing}
    \FeatNode{x}{\off}{\yL}{$X^\ast$}{User Attributes}
    \draw[obsflow] (mg)--(m);
    \draw[obsflow] (m)--(x);
  \end{scope}
  \begin{scope}[shift={(\xa+1.5*\colw,0)}]
    \InsNode{l}{-\off}{\yI}{$I_L$}{consumer\_lag}
    \InsNode{a}{\off}{\yI}{$I_A$}{snapshot\_age}
    \SrcNode{z}{-\off}{\yL}{$Z_H$}{Listening Events}
    \FeatNode{x}{\off}{\yL}{$X^\ast$}{Listening Activity}
    \draw[obsflow] (l)--(z);
    \draw[obsflow] (a)--(x);
    \draw[obsflow] (z)--(x);
  \end{scope}
  \begin{scope}[shift={(\xa+2.5*\colw,0)}]
    \InsNode{v}{\off}{\yI}{$I_V$}{version}
    \SrcNode{z}{-\off}{\yL}{$Z_S$}{Payment Method}
    \FeatNode{x}{\off}{\yL}{$X^\ast$}{Payment Type}
    \draw[obsflow] (v)--(x);
    \draw[obsflow] (z)--(x);
  \end{scope}
\end{tikzpicture}}

\tikzset{
  trhead/.style={inner sep=0pt, font=\fontsize{8.35}{9.55}\selectfont},
  trmode/.style={inner sep=0pt, anchor=west, font=\small},
  trgroup/.style={inner sep=0pt, align=center, font=\fontsize{8.3}{9.5}\selectfont\itshape},
  trcell/.style={inner sep=0pt, font=\small}
}
\newcommand{\TrBox}[5]{%
  \path[fill=panelgray,rounded corners=1.7pt] (#1+\trin,#3-\trin) rectangle (#2-\trin,#4+\trin);
  \node[trcell] at ({(#1+#2)/2},{(#3+#4)/2}) {#5};}

\newcommand{\TransportSketch}{%
\begin{tikzpicture}[x=1cm,y=1cm]
  \def\W{13.9}\def\xa{1.72}\def\xb{5.12}\def\xc{7.36}
  \def\rh{.66}\def\gg{.16}\def\trin{.06}
  \def\yH{.40}%
  \pgfmathsetmacro{\yA}{0}                      %
  \pgfmathsetmacro{\yAm}{\yA-\rh}\pgfmathsetmacro{\yAb}{\yA-2*\rh}
  \pgfmathsetmacro{\yB}{\yAb-\gg}
  \pgfmathsetmacro{\yBm}{\yB-\rh}\pgfmathsetmacro{\yBb}{\yB-2*\rh}
  \pgfmathsetmacro{\yC}{\yBb-\gg}
  \pgfmathsetmacro{\yCm}{\yC-\rh}\pgfmathsetmacro{\yCb}{\yC-2*\rh}
  \pgfmathsetmacro{\yEnd}{\yCb-.06}
  \path[use as bounding box] (0,\yEnd) rectangle (\W,\yA+\yH+.06);
  \draw[draw=linemid,line width=\obsrulewidth] (0,\yA+\yH+.06)--(\W,\yA+\yH+.06);
  \draw[draw=linemid,line width=.36pt] (0,\yA+.06)--(\W,\yA+.06);
  \foreach \x in {\xa,\xb,\xc}{\draw[draw=linemid,line width=.36pt] (\x,\yA+.06)--(\x,\yA+\yH+.06);}
  \draw[draw=linemid,line width=.36pt] (\xa,\yEnd)--(\xa,\yA+.06);
  \draw[draw=linemid,line width=\obsrulewidth] (0,\yEnd)--(\W,\yEnd);
  \foreach \y in {\yAb,\yBb}{\draw[draw=linemid,line width=.36pt] (0,\y-.5*\gg)--(\xa,\y-.5*\gg);}
  \node[trhead] at ({(\xa+\xb)/2},\yA+.06+.5*\yH) {Failure Mode};
  \node[trhead] at ({(\xb+\xc)/2},\yA+.06+.5*\yH) {\textbf{Observe}};
  \node[trhead] at ({(\xc+\W)/2},\yA+.06+.5*\yH) {\textbf{Reconstruct}};
  \node[trgroup] at (.5*\xa,{(\yA+\yAb)/2}) {Missingness};
  \node[trgroup] at (.5*\xa,{(\yB+\yBb)/2}) {Freshness};
  \node[trgroup] at (.5*\xa,{(\yC+\yCb)/2}) {Consistency};
  \node[trmode] at (\xa+.16,{(\yA+\yAm)/2})  {M1: Missing field};
  \node[trmode] at (\xa+.16,{(\yAm+\yAb)/2}) {M2: Failed join};
  \node[trmode] at (\xa+.16,{(\yB+\yBm)/2})  {F1: Snapshot staleness};
  \node[trmode] at (\xa+.16,{(\yBm+\yBb)/2}) {F2: Event-stream lag};
  \node[trmode] at (\xa+.16,{(\yC+\yCm)/2})  {C1: Dependency change};
  \node[trmode] at (\xa+.16,{(\yCm+\yCb)/2}) {C2: Workflow update};
  \TrBox{\xb}{\xc}{\yA}{\yAm}{$P(I_M \mid I_J=0)$}
  \TrBox{\xb}{\xc}{\yAm}{\yAb}{$P(I_J)$}
  \TrBox{\xb}{\xc}{\yB}{\yBm}{$P(I_A \mid C)$}
  \TrBox{\xb}{\xc}{\yBm}{\yBb}{$P(I_L \mid C)$}
  \TrBox{\xb}{\xc}{\yC}{\yCm}{$P(X^\ast \mid Z)$}
  \TrBox{\xb}{\xc}{\yCm}{\yCb}{$P(I_V)$}
  \TrBox{\xc}{\W}{\yA}{\yAb}{$\tilde I_{M/J} \sim P_E(I_{M/J}),\quad \hat X_R^\ast = F_{\mathrm{imp}}\!\left(F_{\mathrm{mask}}(X_R^\ast,\tilde I_{M/J})\right)$}
  \TrBox{\xc}{\W}{\yB}{\yBm}{$\tilde I_A \sim P_E(I_A \mid C_R),\quad \hat X_R^\ast = F_{\mathrm{replace}}(X_R^\ast,\tilde I_A)$}
  \TrBox{\xc}{\W}{\yBm}{\yBb}{$\tilde I_L \sim P_E(I_L \mid C_R),\quad \hat X_R^\ast = F_{\mathrm{agg}}\!\left(F_{\mathrm{truncate}}(Z_R,\tilde I_L)\right)$}
  \TrBox{\xc}{\W}{\yC}{\yCm}{$\hat X_R^\ast \sim P_E(X^\ast \mid Z_R)$}
  \TrBox{\xc}{\W}{\yCm}{\yCb}{$\tilde I_V \sim P_E(I_V),\quad \hat X_R^\ast \sim P_E(X^\ast \mid Z_R,\tilde I_V)$}
\end{tikzpicture}}

\newcommand{\gcirc}{\tikz[baseline=-.6ex]{\node[circle,draw=linecharcoal,fill=nodeblue,line width=.3pt,minimum size=2.5mm,inner sep=0pt]{};}}
\newcommand{\garrow}{\tikz[baseline=-.6ex]{\draw[draw=linecharcoal,line width=.35pt,-{Latex[length=1.1mm,width=.8mm]}] (0,0)--(3.6mm,0);}}

\begin{figure*}[b]
  \centering
  \begin{subfigure}[t]{\textwidth}
    \centering
    \ObsSketch
    \caption{\textbf{Observability of the materialization process}. Nodes ( \gcirc\ ) denote recorded variables, arrows ( \garrow\ ) represent dependencies. \textit{Italics} name the variables as in the churn prediction example, for illustration.}
    \label{fig:observability}
  \end{subfigure}

  \par\vspace{1mm}
  \begin{subfigure}[t]{\textwidth}
    \centering
    \TransportSketch
    \caption{\textbf{Materialization Transport}. Observing and reconstructing six canonical failure modes (Section~\ref{sec:transport_template}).}
    \label{tab:patterns}
  \end{subfigure}

  \vspace{-2mm}
  \caption{Observability and transport of materialization changes. Observe compares $P(\cdot)$ between $R$ and $E$; $P_E$ is a distribution in $E$, subscript $R$ a value in $R$, $C$ unaffected context variables (e.g., time features), $\tilde I$ metadata assigned from $E$, $F$ a reconstruction operation, $\hat X_R^\ast$ reconstructed features.}
\end{figure*}

For a feature or group of features materialized by the same operation, we associate the resulting values $X^\ast$ with the relevant \textit{metadata} $I$, \textit{source data} $Z$, or both (Figure~\ref{fig:observability}). This association enables us to \textbf{observe} changes in the materialization mechanism of the corresponding features. \textit{Metadata} $I$ can be used to expose changes in how these operations materialize the data, such as an increase in the proportion of failed joins or in the staleness of the activity snapshots used, by comparing $P(I)$ between $R$ and $E$. When $I$ depends on unaffected context variables $C$, we instead compare $P(I\mid C)$ between $R$ and $E$ at the same values of $C$. \textit{Source data} $Z$, in contrast, allow us to observe changes in how an operation transforms its inputs into feature values. For example, recording payment method IDs alongside the resulting payment types allows us to compare $P(X^\ast\mid Z)$ between $R$ and $E$ and determine whether the same inputs produce different outputs. In either case, a change can be flagged using a divergence threshold or test of independence~\cite{kullback1951information,budhathokik2021why}.

Given an observed mechanism change, we define how to
\textbf{reconstruct} the counterfactual feature values $X^\ast$
that the reference observations would have had if they had been
materialized under the corresponding mechanism behavior observed
in $E$. Where metadata describe the change, we use $P_E(I)$, or
$P_E(I\mid C)$ given the reference context, to determine assignments
to reference observations. The reconstructed $X^\ast$ values are then
produced from the corresponding reference features or source data,
using the assigned metadata where needed. This can be specified by a
deterministic function $F$ when the changed computation can be
executed or replayed directly, or by a probabilistic model $P_E$
when its behavior instead has to be inferred from inputs and outputs
recorded in $E$ (e.g., when the deployed transformation cannot be reproduced directly from its implementation).

These two steps define a general template for declaring
materialization transports. For each relevant failure mode, the user
specifies the affected features, the relevant metadata $I$ and/or
source data $Z$, any conditioning context $C$, and how the affected
features are reconstructed in $R$. The six transport definitions below instantiate this template through different forms of observability and reconstruction for common failure modes, grouped by data quality dimension (Figure~\ref{tab:patterns}).

\noindent\textbf{Missingness.} Missingness increases in individual fields or across a joined feature group, and the pipeline imputes the missing values before prediction~\cite{breck2019data,shankars2024we,papasian2020ml}. To reproduce this change, both transports (M1-M2) assign failures to the corresponding reference features, preserve existing missing values, and apply the same imputation function as the deployed pipeline.

\begin{itemize}
\item[\textbf{M1}] \textit{Missing field}. The missingness rate of a single field increases (e.g., age). We use its \texttt{is\_missing} flag before imputation to estimate the probability of additional missingness, accounting for missingness already present in $R$. For fields obtained through a join, we measure and assign field-level missingness only where the join succeeds.
\item[\textbf{M2}] \textit{Failed join}. A failed join makes all features from the joined table unavailable (e.g., age and country from the user attributes table). We track \texttt{join\_mismatch} and assign additional failures
to successful reference joins to reproduce the rate observed
in $E$, applying each assigned failure to the entire feature
group.
\end{itemize}

\noindent\textbf{Freshness.} A delayed update leaves recent information out of the features~\cite{breck2017ml,shankar2023automatic,wooders2023ralf,zinkevich2017rules}. Both transports reproduce the corresponding delay effect on the reference features.

\begin{itemize}
\item[\textbf{F1}] \textit{Snapshot staleness}. In a batch pipeline that refreshes snapshots on a regular cadence (e.g., daily/weekly), a failed refresh leaves the pipeline serving an older snapshot (e.g., of listening activity aggregates). We track \texttt{snapshot\_age}, the time since the served snapshot was materialized~\cite{feast_metrics_implementation,tecton_materialization_monitoring}. As a stale snapshot persists across scheduled refreshes, its age increases. To capture this variation over the refresh schedule, we sample an age from $P_E(I_A \mid C)$ conditional on day of week for a reference prediction at $t$. We retrieve the historical snapshot with data cutoff $t-\texttt{snapshot\_age}$ and replace the reference features with its stored values.

\item[\textbf{F2}] \textit{Event stream lag}. An online aggregation pipeline falls behind incoming events, leaving recent events out of its aggregates. We record \texttt{consumer\_lag} as the time between prediction and the processing cutoff used to construct the aggregates. Because processing lag can vary systematically over the course of the day, we sample it from $P_E(I_L\mid C)$ conditional on time of day. For a reference prediction at time $t$, we truncate its event history at $t-\texttt{consumer\_lag}$ and recompute the affected features, keeping aggregation windows anchored at $t$.

\end{itemize}

\noindent\textbf{Consistency.} An update or fault in a transformation can change the functional dependency between its source inputs and the resulting feature values, causing the same inputs to produce different values~\cite{breck2019data,schelters2021jenga}. We model these mappings probabilistically to reconstruct the features in $R$ from their source
data, without requiring access to the implementation of the changed
transformation.

\begin{itemize}
\item[\textbf{C1}] \textit{Dependency change}. A source-to-feature mapping changes, such as the lookup from payment method IDs to payment types. We compare the feature values produced for the same source inputs in $R$ and $E$. To reproduce the change, we sample from the relationship estimated in $E$, $P_E(X^\ast \mid Z)$, given the source data values in $R$.

\item[\textbf{C2}] \textit{Workflow update}. A workflow revision changes how a group of features is computed (e.g., several billing
transformations are updated together). During a partial rollout,
different observations can be processed by old and new workflow
versions within the same period. We track \texttt{version} $I_V$ to identify which version of the feature definitions was applied to each
observation, assuming the computations within each version remain
fixed. For each reference observation, we sample a version from its distribution in $E$ and reconstruct the feature group from its reference source inputs using $P_E(X^\ast\mid Z,I_V)$.
\end{itemize}

\subsubsection{Population Transport}
\label{sec:population_transport}

A population transport uses importance weighting (IW)
\cite{sugiyama2007covariate,chen2021mandoline} to reproduce the part
of the population change observable in comparable data. Because
identified materialization changes can also affect the observed
features, we estimate this change only from data whose materialization
remains comparable between $R$ and $E$. We denote by $A$ the variables
used for population weighting. Building on the notion of negative
controls~\cite{lipsitch2010negative}, we can include in $A$ variables
whose computation does not depend on the operations identified as
changed, but whose values can still vary with the population.
Applying IW on comparable variables $A$ then approximately reproduces
the population change in $R$ as captured through those variables,
subject to the assumptions explained below.

Comparability can also be established at the observation level.
When a materialization change provides a row-level indicator
(e.g., \texttt{join\_mismatch} or \texttt{version}), the affected
features remain comparable for observations not exposed to that
change, provided no other identified change affects them.
If affected status is independent of population attributes,
these observations still represent the evaluation population.
We can then include those features in $A$ and estimate IW using
this subset of evaluation observations. The resulting weights
are applied to all reference observations.

In either case, population transport weights the reference
observations according to the distribution of $A$ in $E$.
Weighting by $P_E(A)/P_R(A)$ reproduces their evaluation distribution
while preserving $P_R(X,Y\mid A)$. Representing the full evaluation
population therefore requires that $R$ covers the evaluation values
of $A$ and that $P(X,Y\mid A)$ remains unchanged between $R$ and $E$.
Population differences that change $P(X,Y\mid A)$ are not
reproduced by weighting on $A$. For example, if a stale activity snapshot makes activity
features $X_2$ incomparable, weighting on unaffected user attributes
$X_1$ captures changes in the user mix but not changes in activity
among otherwise similar users, so $X_2$ remains distributed according
to $P_R(X_2\mid X_1)$.

\subsection{Online Model Testing}
\label{sec:testing}

Online model testing composes the available transports on $R$ to
reproduce the conditions under which performance is to be evaluated.
Materialization transports reconstruct the affected features, while
population transport weights the resulting reference observations.
We then run the model on the reconstructed features and score its predictions against the labels in $R$ using the population weights. Table~\ref{tab:testing_estimands} summarizes the resulting estimands.

\begin{table}[b]
  \centering
  \caption{Performance estimands, with $V$ denoting performance and $G_E,M_E$ the current population and materialization states. $[j\leftarrow R]$ indicates bringing back a mechanism $j$ to its reference state.}
  \vspace{-0.7em}
  \label{tab:testing_estimands}
  \begingroup
  \footnotesize
  \definecolor{testingpanelgray}{RGB}{246,246,246}
  \setlength{\tabcolsep}{5pt}
  \renewcommand{\arraystretch}{1.30}
  \renewcommand{\tabularxcolumn}[1]{m{#1}}
  \begin{tabularx}{\linewidth}{@{}%
    >{\raggedright\arraybackslash}m{.22\linewidth}%
    !{\color{black!38}\vrule width .35pt}%
    >{\columncolor{testingpanelgray}[5pt][5pt]}l%
    !{\color{black!38}\vrule width .35pt}%
    >{\raggedright\arraybackslash}X@{}}
    \arrayrulecolor{black}\specialrule{.9pt}{0pt}{2pt}
    \textbf{Estimand}
      & \multicolumn{1}{>{\columncolor{white}[5pt][5pt]}l!{\color{black!38}\vrule width .35pt}}{\textbf{Formalization}}
      & \textbf{Conditions} \\
    \specialrule{.5pt}{2pt}{3pt}
    Current performance (Q1)
      & $V(G_E,M_E)$
      & Current population and materialization. \\
    \arrayrulecolor{black!30}\specialrule{.4pt}{2.6pt}{2.6pt}
    Repair gain $\tau_j$ (Q2)
      & $V(G_E,M_E[j\leftarrow R])-V(G_E,M_E)$
      & Current population; other mechanisms remain current. \\
    \arrayrulecolor{black}\specialrule{.9pt}{3pt}{0pt}
  \end{tabularx}
  \endgroup
\end{table}

\noindent\textbf{Current performance (Q1).} The target $V(G_E,M_E)$ is model performance under the population and materialization conditions in $E$. To estimate it, we apply the transports for all identified materialization changes to $R$ and evaluate the resulting predictions using the population weights estimated from the comparable data.

\noindent\textbf{Repair gain (Q2).} We compare the current performance test with a second test in which only the selected materialization mechanism is restored to its condition in $R$. Both tests therefore reproduce the current population and all other identified materialization changes, since the gain from the selected repair can depend on these conditions (Section~\ref{sec:problem}).

\section{Experimental Setup}
\label{sec:experimental_setup}

We evaluate whether composing population and materialization transports improves performance estimation and whether representing population change improves repair gain estimation. We introduce controlled materialization faults into real-world datasets that already exhibit population changes, following a common approach in data quality research to create semi-synthetic shifts with known fault occurrence and performance effects~\cite{arocena2015messing,abedjan2016detecting,schelters2021jenga,jager2024data,jung2025towards}.

\subsection{Datasets}

\subsubsection{TableShift}\label{sec:tableshift}

We use the twelve publicly available datasets from the TableShift benchmark to evaluate estimates under controlled population changes~\cite{gardner2023tableshift}. For each dataset, the benchmark provides a context variable, such as geographical region or admission source, which gives us a controlled way to vary the population composition between $R$ and $E$. This variable partitions observations into the benchmark's in-domain and out-of-domain populations. We retain the benchmark training and validation partitions and sample disjoint, equally sized reference and evaluation batches from held-out observations. Across five settings, the out-of-domain shares in $R/E$ are $40/60\%$, $30/70\%$, $20/80\%$, $10/90\%$, and $0/100\%$, progressively increasing the difference in population composition. For a geography-based split, for example, this ranges from both batches containing observations from both sets of regions to $R$ and $E$ containing observations from different sets of regions. The context variable is excluded from the model and the estimators compared here.

\subsubsection{Temporal Prediction Cases}\label{sec:temporal}

\noindent\textbf{Churn.} We use the KKBOX churn prediction dataset~\cite{kkbox}, adapting a shared competition solution~\cite{infinitewing_kkbox_solution} to daily invocation. The model scores subscriptions expiring seven days later and predicts failure to renew within 30 days after expiry. To study snapshot staleness (F1), we implement a daily batch aggregation pipeline that computes activity features from logged user activity, such as listening counts and total listening time over recent windows.

\noindent\textbf{Trip.} We use the Kaggle trip duration
dataset~\cite{taxi_Kaggle}, with geographic and calendar features informed by a shared competition solution~\cite{taxi_Kaggle_solution}. We implement an online pipeline computing recency features and aggregates over the preceding hour from trips completed before prediction, including their observed durations. Under simulated lag, only trips completed by the earlier processing cutoff contribute to these features. Although labels become available relatively quickly, we use this dataset to test event stream lag (F2) under temporal population change, complementing the daily snapshots and longer activity windows in \textit{Churn}.

\noindent\textbf{Default.} We use the cleaned LendingClub data for loan default prediction, used in~\cite{ariza2020explainability,sanz2024credit}, containing resolved loans issued from June 2007 to December 2018. We use all structured columns available in the dataset. Identifiers, dates, and text are excluded from the model features and estimators.

We split each temporal dataset chronologically by calendar day (\textit{Churn}, \textit{Trip}) or month (\textit{Default}). The first 30\%, 40\%, or 50\% of these units form the training period, the next 10\% the reference period, and the remainder the evaluation period. From each evaluation period, we sample up to 20 non-overlapping batches: weekly for \textit{Churn}, daily for \textit{Trip}, and monthly for \textit{Default}. For \textit{Default}, we sample only from the first 24 months of the evaluation period (to limit censoring bias, as recently issued loans are included only if they resolved early).

\subsection{Experimental Protocol}

\noindent\textbf{Subject models.} The subject models are the models whose performance we estimate. We use three model families: linear models (logistic regression for classification and ridge regression for \textit{trip}, which is the only regression case), LightGBM, and multilayer perceptrons (MLPs), each tuned via grid search. For the temporal cases, we split the development data chronologically into 80\% for training and 20\% for validation. Performance is measured by F1, accuracy, and ROC-AUC for classification, and MAE, RMSE, and $R^2$ for \textit{trip}, whose target is $\log(1+\mathrm{duration})$ in seconds.

\noindent\textbf{Materialization faults.} For each reference-evaluation comparison, we introduce controlled faults in $E$ and estimate their effects using labeled $R$. We simulate the six failure modes in Section~\ref{sec:transport_template}. Missingness is assigned at random, making one feature unavailable with probability 50\% or 100\% (M1), or a feature group jointly unavailable with probability 25\% or 50\% (M2). For M2, temporal cases use declared feature groups, while TableShift uses sampled groups of up to five features. Consistency faults change category mappings, scales, or clipping thresholds at two severities, affecting one feature throughout $E$ (C1) or a sampled group of two to four features under a shared version flag assigned to 50\% of observations at random (C2).

Freshness faults follow whether a case relies on batch features summarizing activity over days or online features tracking recent events, such as the preceding hour. In \textit{Churn}, the activity snapshot fails to refresh for seven or fourteen days (F1). In \textit{Trip}, online aggregation experiences two daily backlog incidents peaking at 10 or 20 minutes (F2). \textit{Default} lacks the event history needed to reconstruct stale features and is therefore excluded from these scenarios.

\noindent\textbf{Transport implementation.} We assume that each introduced materialization change has been identified and that the observations required by its transport are available. Because the changes are introduced under controlled conditions, this lets us evaluate the transports described in Section~\ref{sec:materialization_transport} while isolating the transport and composition questions in RQ1 and RQ2 from the separate problem of detecting changes in the materialization process. We later relax this assumption by leaving some concurrent materialization changes unobserved and examine how the estimates are affected when changes are not flagged and their transports therefore cannot be included (Section~\ref{sec:experiment_partial}). 

For each scenario, materialization transport $M$ instantiates the corresponding transport described in Section~\ref{sec:materialization_transport} to reproduce the identified change on $R$. Missingness and consistency transports use marginal assignment. Freshness transports sample the corresponding $I$ conditional on the day of week in \textit{Churn} and on the time of day in \textit{Trip}. Population transport $G$ is constructed as described in Section~\ref{sec:population_transport} from the data that remain comparable under that change. In the main experiments, IW uses logistic regression with
regularization strength $C=0.05$ and a weight cap of 5. The
replication package includes a separate sensitivity analysis of
these settings on an earlier experimental design.

\subsection{Use of AI tools}
OpenAI Codex was used throughout exploratory development, debugging, and final implementation of the experimental code. The authors selected and provided the datasets, defined the experimental design and implementation specifications, and reviewed and tested the resulting code. Anthropic Claude Code was used to implement and visually refine designed figures. The values reported in empirical plots and tables were populated programmatically from the experimental outputs.

\section{Experiments and Evaluation}
\label{sec:experiments}

We organize the evaluation around two RQs, each paired with a hypothesis derived from the formulation in Section~\ref{sec:problem}. Sections~\ref{sec:experiment_total} and~\ref{sec:experiment_gain} evaluate current performance and repair gain estimation, respectively. Section~\ref{sec:experiment_partial} then examines sensitivity to partial mechanism observability by varying which concurrent materialization changes are observed.

Throughout the evaluation, $\mathrm{Transport}(G,M)$ denotes the use of both population and materialization transports. $\mathrm{Transport}(G)$ and $\mathrm{Transport}(M)$ retain only the population and materialization transport, respectively. We evaluate all six materialization transports described in Section~\ref{sec:transport_patterns} and aggregate results by data quality dimension, grouping M1-M2 as missingness, F1-F2 as freshness, and C1-C2 as consistency. For each case and performance metric, we report mean absolute estimation errors averaged across the corresponding configurations described in Section~\ref{sec:experimental_setup}.

\subsection{Performance estimation under concurrent population and materialization change}\label{sec:experiment_total}

\noindent\textbf{RQ1.}
When population and materialization changes occur together, does composing their transports improve estimation of current model performance?

\noindent\textbf{H1.}
Population and materialization changes jointly determine the distribution on which the model performs. When both affect performance and can be adequately transported, we expect composing their transports to reduce estimation error relative to transporting either source alone.

\makeatletter
\ifx\documentclass\@twoclasseserror
  \makeatother\let\EOFinish\relax
\else
  \makeatother
  \documentclass[10pt]{article}
  \usepackage[a4paper,textwidth=395.82pt,top=20mm,bottom=20mm]{geometry}
  \IfFileExists{newtxtext.sty}{\usepackage{newtxtext,newtxmath}}{\usepackage{mathptmx}}
  \usepackage{amsmath,array,tabularx,booktabs,colortbl,multirow,xcolor,graphicx,tikz}
  \usepackage[font=small,labelfont=bf]{caption}
  \pagestyle{empty}
  \begin{document}
  \def\EOFinish{\end{document}}
\fi
\providecolor{panelgray}{RGB}{246,246,246}
\providecolor{linemid}{RGB}{72,72,72}
\providecolor{glosstext}{RGB}{70,70,70}
\providecommand{\EOz}{\phantom{0}}
\providecommand{\EOHead}{\scriptsize}
\providecommand{\EOTop}{\arrayrulecolor{black}\specialrule{.9pt}{0pt}{2pt}}
\providecommand{\EOMid}{\arrayrulecolor{black}\specialrule{.5pt}{2pt}{2.6pt}}
\providecommand{\EOGroup}{\arrayrulecolor{black!30}\specialrule{.4pt}{2.6pt}{2.6pt}\arrayrulecolor{black}}
\providecommand{\EOBottom}{\arrayrulecolor{black}\specialrule{.9pt}{3pt}{0pt}}
\providecommand{\EOBand}{\cellcolor{panelgray}}

\providecolor{nodegreen}{RGB}{217,242,208}
\providecolor{nodeblue}{RGB}{205,243,255}
\providecolor{linecharcoal}{RGB}{48,48,48}
\providecolor{gInk}{HTML}{3A9F2D}
\providecolor{mInk}{HTML}{1B8FBF}
\providecommand{\MechName}[1]{\boldsymbol{#1}}
\tikzset{
  eochip/.style    ={rounded corners=1.3pt,inner xsep=1.6pt,inner ysep=0.9pt,minimum width=3.7mm,font=\scriptsize},
  eochip on/.style ={eochip,draw=linecharcoal,line width=.33pt},
  eochip off/.style={eochip,draw=linemid,fill=none,text=black!40,line width=.24pt,dash pattern=on 1.1pt off .9pt},
}
\ifdefined\EOChipG\else\DeclareRobustCommand{\EOChipG}[1]{\ifnum#1=1\relax\tikz[baseline=(c.base)]{\node[eochip on,fill=nodegreen](c){$\MechName{G}$};}\else\tikz[baseline=(c.base)]{\node[eochip off](c){$\MechName{G}$};}\fi}\fi
\ifdefined\EOChipM\else\DeclareRobustCommand{\EOChipM}[1]{\ifnum#1=1\relax\tikz[baseline=(c.base)]{\node[eochip on,fill=nodeblue](c){$\MechName{M}$};}\else\tikz[baseline=(c.base)]{\node[eochip off](c){$\MechName{M}$};}\fi}\fi
\ifdefined\EOChips\else\DeclareRobustCommand{\EOChips}[2]{\mbox{\EOChipG{#1}\hspace{1.2pt}\EOChipM{#2}}}\fi
\ifdefined\EODot\else\DeclareRobustCommand{\EODot}[2]{\ifnum#2=1\relax\tikz[baseline=-.6ex]{\fill[#1] (0,0) circle (1.7pt);}\else\tikz[baseline=-.6ex]{\draw[black!35,line width=.45pt] (0,0) circle (1.5pt);}\fi}\fi

\begin{table}[b]
  \caption{\textbf{Estimation error of current performance} (Mean absolute error x100). Transport tests reproduce the population change ( \EOChipG{1} ), the materialization faults ( \EOChipM{1} ); a dashed box marks a mechanism that is not transported. Naive reports the reference performance unchanged. Bold marks the lowest per scenario (best).}
  \label{tab:e1-overview}
  \begingroup
  \footnotesize
  \setlength{\tabcolsep}{2.2pt}
  \renewcommand{\arraystretch}{1.13}
\begin{tabularx}{\linewidth}{@{}>{\centering\arraybackslash}p{15mm}!{\color{black!38}\vrule width .35pt}>{\centering\arraybackslash}p{2.6mm}>{\raggedright\arraybackslash}p{9.6mm}!{\color{black!38}\vrule width .35pt}>{\hsize=0.9727\hsize\centering\arraybackslash}X>{\hsize=0.9727\hsize\centering\arraybackslash}X>{\hsize=0.9727\hsize\centering\arraybackslash}X!{\color{black!38}\vrule width .35pt}>{\hsize=0.9727\hsize\centering\arraybackslash}X>{\hsize=0.9727\hsize\centering\arraybackslash}X>{\hsize=0.9727\hsize\centering\arraybackslash}X!{\color{black!38}\vrule width .35pt}>{\hsize=0.9727\hsize\centering\arraybackslash}X>{\hsize=0.9727\hsize\centering\arraybackslash}X>{\hsize=1.3000\hsize\centering\arraybackslash}X!{\color{black!38}\vrule width .35pt}>{\hsize=0.9727\hsize\centering\arraybackslash}X>{\hsize=0.9727\hsize\centering\arraybackslash}X>{\hsize=0.9727\hsize\centering\arraybackslash}X@{}}
\EOTop
\multicolumn{3}{@{}l!{\color{black!38}\vrule width .35pt}}{Estimation error $\downarrow$} & \multicolumn{3}{c!{\color{black!38}\vrule width .35pt}}{\textbf{TableShift}} & \multicolumn{3}{c!{\color{black!38}\vrule width .35pt}}{\textbf{Churn}} & \multicolumn{3}{c!{\color{black!38}\vrule width .35pt}}{\textbf{Trip}} & \multicolumn{3}{c}{\textbf{Default}} \\
\arrayrulecolor{linemid}\cmidrule[.4pt](l{3pt}r{3pt}){4-6}\cmidrule[.4pt](l{3pt}r{3pt}){7-9}\cmidrule[.4pt](l{3pt}r{3pt}){10-12}\cmidrule[.4pt](l{3pt}r{0pt}){13-15}\arrayrulecolor{black}
\multicolumn{1}{c!{\color{black!38}\vrule width .35pt}}{\textbf{Scenario}} & \multicolumn{2}{l!{\color{black!38}\vrule width .35pt}}{\textbf{Estimator}} & \makebox[0pt]{\EOHead F1} & \makebox[0pt]{\EOHead Acc} & \makebox[0pt]{\EOHead AUC} & \makebox[0pt]{\EOHead F1} & \makebox[0pt]{\EOHead Acc} & \makebox[0pt]{\EOHead AUC} & \makebox[0pt]{\EOHead MAE} & \makebox[0pt]{\EOHead RMSE} & \makebox[0pt]{\EOHead $R^2$} & \makebox[0pt]{\EOHead F1} & \makebox[0pt]{\EOHead Acc} & \makebox[0pt]{\EOHead AUC} \\
\EOMid
 & \multicolumn{1}{>{\columncolor{panelgray}[\dimexpr\tabcolsep-.35pt\relax][\tabcolsep]\centering\arraybackslash}p{2.6mm}}{} & \EOBand Naive & \EOBand \EOz4.81 & \EOBand 3.38 & \EOBand 2.96 & \EOBand 11.21 & \EOBand 13.45 & \EOBand 15.08 & \EOBand \EOz3.82 & \EOBand \EOz5.13 & \EOBand \EOz7.59 & \EOBand \textbf{2.90} & \EOBand 6.71 & \multicolumn{1}{>{\columncolor{panelgray}[\tabcolsep][0pt]}c@{}}{2.80} \\
 & \multicolumn{1}{>{\columncolor{panelgray}[\dimexpr\tabcolsep-.35pt\relax][\tabcolsep]\centering\arraybackslash}p{2.6mm}}{} & \EOBand IW & \EOBand \EOz3.99 & \EOBand 2.55 & \EOBand 2.83 & \EOBand \EOz9.87 & \EOBand \EOz4.87 & \EOBand \EOz7.06 & \EOBand \EOz3.58 & \EOBand \EOz4.96 & \EOBand \EOz7.49 & \EOBand 3.45 & \EOBand 1.80 & \multicolumn{1}{>{\columncolor{panelgray}[\tabcolsep][0pt]}c@{}}{2.12} \\
 & \multicolumn{1}{>{\columncolor{panelgray}[\dimexpr\tabcolsep-.35pt\relax][\tabcolsep]\centering\arraybackslash}p{2.6mm}}{\multirow{-3}{*}{\rotatebox[origin=c]{90}{\fontsize{6.6pt}{7pt}\selectfont \textcolor{glosstext}{Baselines}}}} & \EOBand OT & \EOBand 12.22 & \EOBand 5.79 & \EOBand 8.70 & \EOBand 15.71 & \EOBand \EOz6.27 & \EOBand 15.46 & \EOBand 12.70 & \EOBand 14.93 & \EOBand 25.72 & \EOBand 4.22 & \EOBand 1.61 & \multicolumn{1}{>{\columncolor{panelgray}[\tabcolsep][0pt]}c@{}}{2.76} \\
\arrayrulecolor{black!45}\cline{2-15}\arrayrulecolor{black}
 &  & \EOChips{1}{0} & \EOz3.80 & 2.54 & 2.83 & \EOz9.51 & \EOz4.50 & \EOz6.86 & \EOz3.57 & \EOz4.95 & \EOz7.48 & 3.29 & 3.00 & 1.88 \\
 &  & \EOChips{0}{1} & \EOz3.95 & 2.69 & 1.93 & 10.23 & 12.32 & 14.14 & \EOz1.42 & \EOz1.73 & \EOz1.35 & 3.08 & 5.60 & 1.95 \\
\multirow{-6}{*}{\textit{Missingness}} & \multirow{-3}{*}{\rotatebox[origin=c]{90}{\fontsize{6.6pt}{7pt}\selectfont \textbf{Transport}}} & \EOChips{1}{1} & \EOz\textbf{2.63} & \textbf{1.70} & \textbf{1.86} & \EOz\textbf{6.84} & \EOz\textbf{2.95} & \EOz\textbf{5.47} & \EOz\textbf{1.18} & \EOz\textbf{1.62} & \EOz\textbf{1.28} & 3.39 & \textbf{1.46} & \textbf{1.15} \\
\arrayrulecolor{black!70}\specialrule{.65pt}{2.6pt}{2.6pt}\arrayrulecolor{black}
 & \multicolumn{1}{>{\columncolor{panelgray}[\dimexpr\tabcolsep-.35pt\relax][\tabcolsep]\centering\arraybackslash}p{2.6mm}}{} & \EOBand Naive & \multicolumn{3}{c!{\color{black!38}\vrule width .35pt}}{} & \EOBand 11.37 & \EOBand 12.58 & \EOBand 14.51 & \EOBand \EOz1.67 & \EOBand \EOz2.05 & \EOBand \EOz1.71 & \multicolumn{3}{c}{} \\
 & \multicolumn{1}{>{\columncolor{panelgray}[\dimexpr\tabcolsep-.35pt\relax][\tabcolsep]\centering\arraybackslash}p{2.6mm}}{} & \EOBand IW & \multicolumn{3}{c!{\color{black!38}\vrule width .35pt}}{} & \EOBand 16.96 & \EOBand \EOz4.96 & \EOBand \EOz\textbf{6.43} & \EOBand \EOz1.54 & \EOBand \EOz2.03 & \EOBand \EOz1.71 & \multicolumn{3}{c}{} \\
 & \multicolumn{1}{>{\columncolor{panelgray}[\dimexpr\tabcolsep-.35pt\relax][\tabcolsep]\centering\arraybackslash}p{2.6mm}}{\multirow{-3}{*}{\rotatebox[origin=c]{90}{\fontsize{6.6pt}{7pt}\selectfont \textcolor{glosstext}{Baselines}}}} & \EOBand OT & \multicolumn{3}{c!{\color{black!38}\vrule width .35pt}}{} & \EOBand 16.22 & \EOBand \EOz6.40 & \EOBand 16.20 & \EOBand 13.40 & \EOBand 15.94 & \EOBand 26.98 & \multicolumn{3}{c}{} \\
\arrayrulecolor{black!45}\cline{2-3}\cline{7-12}\arrayrulecolor{black}
 &  & \EOChips{1}{0} & \multicolumn{3}{c!{\color{black!38}\vrule width .35pt}}{} & \EOz9.02 & \EOz4.12 & \EOz6.58 & \EOz1.49 & \EOz1.97 & \EOz1.59 & \multicolumn{3}{c}{} \\
 &  & \EOChips{0}{1} & \multicolumn{3}{c!{\color{black!38}\vrule width .35pt}}{} & 11.57 & 12.66 & 14.59 & \EOz1.46 & \EOz1.82 & \EOz1.48 & \multicolumn{3}{c}{} \\
\multirow{-6}{*}{\textit{Freshness}} & \multirow{-3}{*}{\rotatebox[origin=c]{90}{\fontsize{6.6pt}{7pt}\selectfont \textbf{Transport}}} & \EOChips{1}{1} & \multicolumn{3}{c!{\color{black!38}\vrule width .35pt}}{\multirow{-6}{*}{\textcolor{black!45}{\normalsize ---}}} & \EOz\textbf{8.46} & \EOz\textbf{3.82} & \EOz6.55 & \EOz\textbf{1.30} & \EOz\textbf{1.77} & \EOz\textbf{1.37} & \multicolumn{3}{c}{\multirow{-6}{*}{\textcolor{black!45}{\normalsize ---}}} \\
\arrayrulecolor{black!70}\specialrule{.65pt}{2.6pt}{2.6pt}\arrayrulecolor{black}
 & \multicolumn{1}{>{\columncolor{panelgray}[\dimexpr\tabcolsep-.35pt\relax][\tabcolsep]\centering\arraybackslash}p{2.6mm}}{} & \EOBand Naive & \EOBand \EOz4.53 & \EOBand 3.32 & \EOBand 2.64 & \EOBand 11.30 & \EOBand 13.51 & \EOBand 15.22 & \EOBand 13.09 & \EOBand 15.52 & \EOBand 70.11 & \EOBand 3.33 & \EOBand 8.63 & \multicolumn{1}{>{\columncolor{panelgray}[\tabcolsep][0pt]}c@{}}{2.93} \\
 & \multicolumn{1}{>{\columncolor{panelgray}[\dimexpr\tabcolsep-.35pt\relax][\tabcolsep]\centering\arraybackslash}p{2.6mm}}{} & \EOBand IW & \EOBand \EOz3.71 & \EOBand 2.42 & \EOBand 2.50 & \EOBand 10.18 & \EOBand \EOz3.90 & \EOBand \EOz6.91 & \EOBand 12.83 & \EOBand 15.30 & \EOBand 70.06 & \EOBand 3.48 & \EOBand 2.80 & \multicolumn{1}{>{\columncolor{panelgray}[\tabcolsep][0pt]}c@{}}{2.06} \\
 & \multicolumn{1}{>{\columncolor{panelgray}[\dimexpr\tabcolsep-.35pt\relax][\tabcolsep]\centering\arraybackslash}p{2.6mm}}{\multirow{-3}{*}{\rotatebox[origin=c]{90}{\fontsize{6.6pt}{7pt}\selectfont \textcolor{glosstext}{Baselines}}}} & \EOBand OT & \EOBand 12.28 & \EOBand 5.88 & \EOBand 9.11 & \EOBand 16.74 & \EOBand \EOz6.35 & \EOBand 16.07 & \EOBand 12.65 & \EOBand 15.12 & \EOBand 34.36 & \EOBand 4.64 & \EOBand 2.05 & \multicolumn{1}{>{\columncolor{panelgray}[\tabcolsep][0pt]}c@{}}{3.33} \\
\arrayrulecolor{black!45}\cline{2-15}\arrayrulecolor{black}
 &  & \EOChips{1}{0} & \EOz3.62 & 2.39 & 2.54 & \EOz8.32 & \EOz3.70 & \EOz6.54 & 12.85 & 15.34 & 69.93 & 3.77 & 5.65 & 2.04 \\
 &  & \EOChips{0}{1} & \EOz4.00 & 2.66 & 2.06 & 11.28 & 13.18 & 15.07 & \EOz1.99 & \EOz2.46 & \EOz7.04 & \textbf{3.18} & 5.44 & 1.79 \\
\multirow{-6}{*}{\textit{Consistency}} & \multirow{-3}{*}{\rotatebox[origin=c]{90}{\fontsize{6.6pt}{7pt}\selectfont \textbf{Transport}}} & \EOChips{1}{1} & \EOz\textbf{2.70} & \textbf{1.62} & \textbf{1.86} & \EOz\textbf{7.56} & \EOz\textbf{3.27} & \EOz\textbf{5.98} & \EOz\textbf{1.37} & \EOz\textbf{1.90} & \EOz\textbf{4.52} & 3.41 & \textbf{1.44} & \textbf{1.10} \\
\EOBottom
\end{tabularx}%
  \endgroup
\end{table}
\EOFinish

\noindent\textbf{Comparison.}
We estimate current model performance on $E$ under concurrent
population and materialization change. We compare $\mathrm{Transport}(G,M)$ with \textit{Naive}, which reports reference performance unchanged; \textit{IW}, which weights reference performance using a domain classifier fitted on reference and served evaluation features; and \textit{OT}, which transfers reference labels to evaluation observations through optimal transport~\cite{decker2024explanatory}.

Two ablations assess the contribution of each transport component. $\mathrm{Transport}(G)$ retains the composed estimator's population transport without changing reference features, while $\mathrm{Transport}(M)$ applies the materialization transport procedure without accounting for population change. We assess all estimates against measured performance on the faulty evaluation data.

\noindent\textbf{Results.}
Table~\ref{tab:e1-overview} shows that composing population and materialization transports generally gives the lowest estimation error across datasets and fault families. The comparison with the ablations supports H1, showing that reproducing either source alone can leave relevant changes in the evaluation conditions unrepresented. Other estimators have slightly lower AUC error in Churn under freshness faults. For Default F1, Naive performs best under missingness while materialization transport alone performs best under consistency faults.

\subsection{Repair gain estimation under population change}\label{sec:experiment_gain}

\input{sections/figures/e21_figure}

\noindent\textbf{RQ2.}
Does transporting population change improve estimation of repair gain in the evaluation population?

\noindent\textbf{H2.}
The repair gain depends on the population in which it is tested. When population change redistributes observations with different responses to a repair, we expect transporting that change to reduce repair gain estimation error relative to testing on the reference population, provided the transport captures the relevant difference.

\noindent\textbf{Comparison.}
Using the same fault cases, we now estimate the performance gain from repairing the introduced fault. Both $\mathrm{Transport}(G,M)$ and $\mathrm{Transport}(M)$ compare faulty and repaired states constructed on $R$. The former applies population transport to both states to estimate the gain in the evaluation population, while the latter makes the comparison in the reference population. \textit{No gain} predicts zero improvement.

We measure the actual gain by comparing performance on the same observations in $E$ with and without the introduced fault. Across the TableShift mixtures, we examine how the benefit of population transport varies with population separation. An additional retrospective diagnostic, $\mathrm{Transport}(G,M)+\mathrm{labels}$, adds $Y$ to the comparable inputs used for population transport using the same materialization transport procedure. This assesses whether population information unavailable before evaluation labels arrive improves estimation.

\noindent\textbf{Results.}
Figure~\ref{fig:e21-changes} shows lower mean repair gain
estimation error with population transport in every reported
dataset-fault-metric combination. This supports H2, showing that testing the same repair on the reference population can misrepresent its benefit under current conditions. The composed estimator also improves on No gain, except for Churn AUC under freshness faults.

Figure~\ref{fig:e21-mixtures} examines this comparison as the TableShift populations differ more in composition. For F1 and accuracy, estimation error increases with population separation. Population transport lowers error at every mixture but does not remove this increasing pattern. Adding labels lowers these errors further, with the additional reduction increasing as the populations separate. This is consistent with the limitation described in Section~\ref{sec:population_transport}: comparable inputs may leave population differences relevant to the repair gain unrepresented. For AUC, estimation error also increases with population separation, while the reduction from population transport
grows across the mixtures. Adding labels does not consistently improve AUC, so the benefit of this additional population information also depends on the performance metric being estimated.

\subsection{Partially identified faults}\label{sec:experiment_partial}

\makeatletter
\ifx\documentclass\@twoclasseserror
  \makeatother\let\EOFinish\relax
\else
  \makeatother
  \documentclass[10pt]{article}
  \usepackage[a4paper,textwidth=395.82pt,top=20mm,bottom=20mm]{geometry}
  \IfFileExists{newtxtext.sty}{\usepackage{newtxtext,newtxmath}}{\usepackage{mathptmx}}
  \usepackage{amsmath,array,tabularx,booktabs,colortbl,multirow,xcolor,graphicx,tikz}
  \usepackage[font=small,labelfont=bf]{caption}
  \usepackage{subcaption}
  \pagestyle{empty}
  \begin{document}
  \def\EOFinish{\end{document}}
\fi
\providecolor{panelgray}{RGB}{246,246,246}
\providecolor{linemid}{RGB}{72,72,72}
\providecolor{glosstext}{RGB}{70,70,70}
\providecommand{\EOz}{\phantom{0}}
\providecommand{\EOHead}{\scriptsize}
\providecommand{\EOTop}{\arrayrulecolor{black}\specialrule{.9pt}{0pt}{2pt}}
\providecommand{\EOMid}{\arrayrulecolor{black}\specialrule{.5pt}{2pt}{2.6pt}}
\providecommand{\EOGroup}{\arrayrulecolor{black!30}\specialrule{.4pt}{2.6pt}{2.6pt}\arrayrulecolor{black}}
\providecommand{\EOBottom}{\arrayrulecolor{black}\specialrule{.9pt}{3pt}{0pt}}
\providecommand{\EOBand}{\cellcolor{panelgray}}

\providecolor{nodegreen}{RGB}{217,242,208}
\providecolor{nodeblue}{RGB}{205,243,255}
\providecolor{linecharcoal}{RGB}{48,48,48}
\providecolor{gInk}{HTML}{3A9F2D}
\providecolor{mInk}{HTML}{1B8FBF}
\providecommand{\MechName}[1]{\boldsymbol{#1}}
\tikzset{
  eochip/.style    ={rounded corners=1.3pt,inner xsep=1.6pt,inner ysep=0.9pt,minimum width=3.7mm,font=\scriptsize},
  eochip on/.style ={eochip,draw=linecharcoal,line width=.33pt},
  eochip off/.style={eochip,draw=linemid,fill=none,text=black!40,line width=.24pt,dash pattern=on 1.1pt off .9pt},
}
\ifdefined\EOChipG\else\DeclareRobustCommand{\EOChipG}[1]{\ifnum#1=1\relax\tikz[baseline=(c.base)]{\node[eochip on,fill=nodegreen](c){$\MechName{G}$};}\else\tikz[baseline=(c.base)]{\node[eochip off](c){$\MechName{G}$};}\fi}\fi
\ifdefined\EOChipM\else\DeclareRobustCommand{\EOChipM}[1]{\ifnum#1=1\relax\tikz[baseline=(c.base)]{\node[eochip on,fill=nodeblue](c){$\MechName{M}$};}\else\tikz[baseline=(c.base)]{\node[eochip off](c){$\MechName{M}$};}\fi}\fi
\ifdefined\EOChips\else\DeclareRobustCommand{\EOChips}[2]{\mbox{\EOChipG{#1}\hspace{1.2pt}\EOChipM{#2}}}\fi
\ifdefined\EODot\else\DeclareRobustCommand{\EODot}[2]{\ifnum#2=1\relax\tikz[baseline=-.6ex]{\fill[#1] (0,0) circle (1.7pt);}\else\tikz[baseline=-.6ex]{\draw[black!35,line width=.45pt] (0,0) circle (1.5pt);}\fi}\fi

\providecolor{overviewGuide}{HTML}{303030}
\providecolor{diaggray}{RGB}{140,140,140}
\providecolor{mixReplay}{HTML}{1B8FBF}
\providecolor{mixComposed}{HTML}{082C3A}
\providecolor{mixDiag}{RGB}{150,150,150}
\providecolor{mixBase}{RGB}{234,234,234}
\tikzset{
  mixbase/.style ={line cap=butt,line join=miter,every node/.style={inner sep=0pt,outer sep=0pt}},
  mixtitle/.style={font=\footnotesize,anchor=base},
  mixtick/.style ={font=\scriptsize},
  mixaxis/.style ={draw=overviewGuide,line width=.6pt},
  pics/eocirc/.style={code={\path[pic actions] (0,0) circle (.72);}},
  pics/eotri/.style ={code={\path[pic actions] (0,.98) -- (-.85,-.49) -- (.85,-.49) -- cycle;}},
  pics/eodiam/.style={code={\path[pic actions] (0,.95) -- (.8,0) -- (0,-.95) -- (-.8,0) -- cycle;}},
}
\ifdefined\EOMixMark\else
\newcommand{\EOMixMark}[6]{%
  \draw[#1,line width=.5pt] (#3,#5) -- (#3,#6);
  \draw[#1,line width=.5pt] (#3-.45,#5) -- (#3+.45,#5);
  \draw[#1,line width=.5pt] (#3-.45,#6) -- (#3+.45,#6);
  \pic[fill=#1,draw=white,line width=.3pt] at (#3,#4) {#2};}
\fi

\providecolor{baseInk}{RGB}{120,120,120}
\providecolor{monoInk}{RGB}{38,38,38}
\tikzset{pics/eosq/.style={code={\path[pic actions] (-.64,-.64) rectangle (.64,.64);}}}
\ifdefined\EOLineRep\else\DeclareRobustCommand{\EOLineRep}{\tikz[x=1mm,y=1mm,baseline=-.6ex]{\draw[monoInk,line width=.9pt] (0,0) -- (5.2,0);\pic[fill=white,draw=monoInk,line width=.55pt,scale=1.05] at (2.6,0) {eosq};}}\fi
\ifdefined\EOLineComp\else\DeclareRobustCommand{\EOLineComp}{\tikz[x=1mm,y=1mm,baseline=-.6ex]{\draw[monoInk,line width=.9pt] (0,0) -- (5.2,0);\pic[fill=monoInk,draw=white,line width=.3pt,scale=1.15] at (2.6,0) {eocirc};}}\fi
\ifdefined\EOpd\else\newdimen\EOpd\fi
\providecommand{\EOFigRule}[2]{\par\EOpd=\prevdepth
  \vskip\dimexpr0pt-(#2)-(#1)\relax\hrule height #1\relax\vskip\dimexpr#2\relax
  \prevdepth=\EOpd}
\begin{figure}[b]
  \centering
  \EOFigRule{.9pt}{2pt}%
  \begin{subfigure}[t]{.493\linewidth}
    \centering
\begin{tikzpicture}[x=1mm,y=1mm,mixbase]
\path[use as bounding box] (0,1.4) rectangle (68.6,36.4);
\node[mixtitle] at (35.80,34.20) {Current performance};
\fill[panelgray] (11.00,29.182) rectangle (60.60,31.00);
\draw[black!10,line width=.35pt] (11.00,11.000) -- (60.60,11.000);
\draw[mixaxis,line width=.45pt] (10.10,11.000) -- (11.00,11.000);
\node[mixtick,anchor=east] at (9.60,11.000) {0};
\draw[black!10,line width=.35pt] (11.00,15.545) -- (60.60,15.545);
\draw[mixaxis,line width=.45pt] (10.10,15.545) -- (11.00,15.545);
\node[mixtick,anchor=east] at (9.60,15.545) {25};
\draw[black!10,line width=.35pt] (11.00,20.091) -- (60.60,20.091);
\draw[mixaxis,line width=.45pt] (10.10,20.091) -- (11.00,20.091);
\node[mixtick,anchor=east] at (9.60,20.091) {50};
\draw[black!10,line width=.35pt] (11.00,24.636) -- (60.60,24.636);
\draw[mixaxis,line width=.45pt] (10.10,24.636) -- (11.00,24.636);
\node[mixtick,anchor=east] at (9.60,24.636) {75};
\draw[black!10,line width=.35pt] (11.00,29.182) -- (60.60,29.182);
\draw[mixaxis,line width=.45pt] (10.10,29.182) -- (11.00,29.182);
\node[mixtick,anchor=east] at (9.60,29.182) {100};
\draw[baseInk,line width=.55pt] (11.00,29.182) -- (60.60,29.182);
\node[mixtick,anchor=west,font=\tiny,text=black!60] at (61.20,29.182) {Naive};
\draw[baseInk,line width=.55pt,dash pattern=on 1.8pt off 1.2pt] (11.00,26.837) -- (60.60,26.837);
\node[mixtick,anchor=west,font=\tiny,text=black!60] at (61.20,26.837) {IW};
\draw[baseInk,line width=.6pt,dash pattern=on .5pt off .9pt] (11.00,24.770) -- (60.60,24.770);
\node[mixtick,anchor=west,font=\tiny,text=black!60] at (61.20,24.770) {OT};
\draw[monoInk,line width=.9pt] (14.00,29.182) -- (24.90,26.716) -- (35.80,24.271) -- (46.70,21.864) -- (57.60,19.499);
\pic[fill=white,draw=monoInk,line width=.55pt,scale=1.05] at (14.00,29.182) {eosq};
\pic[fill=white,draw=monoInk,line width=.55pt,scale=1.05] at (24.90,26.716) {eosq};
\pic[fill=white,draw=monoInk,line width=.55pt,scale=1.05] at (35.80,24.271) {eosq};
\pic[fill=white,draw=monoInk,line width=.55pt,scale=1.05] at (46.70,21.864) {eosq};
\pic[fill=white,draw=monoInk,line width=.55pt,scale=1.05] at (57.60,19.499) {eosq};
\draw[monoInk,line width=.9pt] (14.00,26.837) -- (24.90,24.285) -- (35.80,21.716) -- (46.70,19.126) -- (57.60,16.458);
\pic[fill=monoInk,draw=white,line width=.3pt,scale=1.15] at (14.00,26.837) {eocirc};
\pic[fill=monoInk,draw=white,line width=.3pt,scale=1.15] at (24.90,24.285) {eocirc};
\pic[fill=monoInk,draw=white,line width=.3pt,scale=1.15] at (35.80,21.716) {eocirc};
\pic[fill=monoInk,draw=white,line width=.3pt,scale=1.15] at (46.70,19.126) {eocirc};
\pic[fill=monoInk,draw=white,line width=.3pt,scale=1.15] at (57.60,16.458) {eocirc};
\draw[mixaxis] (11.00,31.00) -- (11.00,11.00) -- (60.60,11.00);
\draw[mixaxis,line width=.45pt] (14.00,11.000) -- (14.00,10.10);
\node[mixtick] at (14.00,8.00) {0};
\draw[mixaxis,line width=.45pt] (24.90,11.000) -- (24.90,10.10);
\node[mixtick] at (24.90,8.00) {1};
\draw[mixaxis,line width=.45pt] (35.80,11.000) -- (35.80,10.10);
\node[mixtick] at (35.80,8.00) {2};
\draw[mixaxis,line width=.45pt] (46.70,11.000) -- (46.70,10.10);
\node[mixtick] at (46.70,8.00) {3};
\draw[mixaxis,line width=.45pt] (57.60,11.000) -- (57.60,10.10);
\node[mixtick] at (57.60,8.00) {4};
\node[mixtick,text=glosstext] at (35.80,3.60) {Materialization changes identified (of four)};
\node[mixtitle,rotate=90,font=\scriptsize,anchor=center] at (1.60,21.00) {Error (\% of Naive)};
\end{tikzpicture}%
    \caption{\textbf{Current performance}, relative to Naive.}
    \label{fig:multifault-performance}
  \end{subfigure}\hfill
  \begin{subfigure}[t]{.493\linewidth}
    \centering
\begin{tikzpicture}[x=1mm,y=1mm,mixbase]
\path[use as bounding box] (0,1.4) rectangle (68.6,36.4);
\node[mixtitle] at (35.80,34.20) {Repair gain};
\draw[black!10,line width=.35pt] (11.00,11.000) -- (60.60,11.000);
\draw[mixaxis,line width=.45pt] (10.10,11.000) -- (11.00,11.000);
\node[mixtick,anchor=east] at (9.60,11.000) {0};
\draw[black!10,line width=.35pt] (11.00,16.000) -- (60.60,16.000);
\draw[mixaxis,line width=.45pt] (10.10,16.000) -- (11.00,16.000);
\node[mixtick,anchor=east] at (9.60,16.000) {20};
\draw[black!10,line width=.35pt] (11.00,21.000) -- (60.60,21.000);
\draw[mixaxis,line width=.45pt] (10.10,21.000) -- (11.00,21.000);
\node[mixtick,anchor=east] at (9.60,21.000) {40};
\draw[black!10,line width=.35pt] (11.00,26.000) -- (60.60,26.000);
\draw[mixaxis,line width=.45pt] (10.10,26.000) -- (11.00,26.000);
\node[mixtick,anchor=east] at (9.60,26.000) {60};
\draw[black!10,line width=.35pt] (11.00,31.000) -- (60.60,31.000);
\draw[mixaxis,line width=.45pt] (10.10,31.000) -- (11.00,31.000);
\node[mixtick,anchor=east] at (9.60,31.000) {80};
\draw[monoInk,line width=.9pt] (14.00,25.466) -- (28.53,24.013) -- (43.07,22.439) -- (57.60,20.705);
\pic[fill=white,draw=monoInk,line width=.55pt,scale=1.05] at (14.00,25.466) {eosq};
\pic[fill=white,draw=monoInk,line width=.55pt,scale=1.05] at (28.53,24.013) {eosq};
\pic[fill=white,draw=monoInk,line width=.55pt,scale=1.05] at (43.07,22.439) {eosq};
\pic[fill=white,draw=monoInk,line width=.55pt,scale=1.05] at (57.60,20.705) {eosq};
\draw[monoInk,line width=.9pt] (14.00,23.397) -- (28.53,21.775) -- (43.07,19.815) -- (57.60,17.579);
\pic[fill=monoInk,draw=white,line width=.3pt,scale=1.15] at (14.00,23.397) {eocirc};
\pic[fill=monoInk,draw=white,line width=.3pt,scale=1.15] at (28.53,21.775) {eocirc};
\pic[fill=monoInk,draw=white,line width=.3pt,scale=1.15] at (43.07,19.815) {eocirc};
\pic[fill=monoInk,draw=white,line width=.3pt,scale=1.15] at (57.60,17.579) {eocirc};
\draw[mixaxis] (11.00,31.00) -- (11.00,11.00) -- (60.60,11.00);
\draw[mixaxis,line width=.45pt] (14.00,11.000) -- (14.00,10.10);
\node[mixtick] at (14.00,8.00) {0};
\draw[mixaxis,line width=.45pt] (28.53,11.000) -- (28.53,10.10);
\node[mixtick] at (28.53,8.00) {1};
\draw[mixaxis,line width=.45pt] (43.07,11.000) -- (43.07,10.10);
\node[mixtick] at (43.07,8.00) {2};
\draw[mixaxis,line width=.45pt] (57.60,11.000) -- (57.60,10.10);
\node[mixtick] at (57.60,8.00) {3};
\node[mixtick,text=glosstext] at (35.80,3.60) {Other materialization changes identified (of three)};
\node[mixtitle,rotate=90,font=\scriptsize,anchor=center] at (1.60,21.00) {Error (\% of No gain)};
\end{tikzpicture}%
    \caption{\textbf{Repair gain}, relative to No gain.}
    \label{fig:multifault-repair}
  \end{subfigure}
  \EOFigRule{.5pt}{-5.3pt}%
  \caption{Estimation error with four concurrent materialization changes, of which only some are identified. Transport tests reproduce these changes alone \mbox{(\,\EOLineRep\,\EOChips{0}{1}\,)} or together with the population change \mbox{(\,\EOLineComp\,\EOChips{1}{1}\,)}. Unidentified materialization changes remain in the data used to fit the population weights.}
  \label{fig:multifault}
\end{figure}
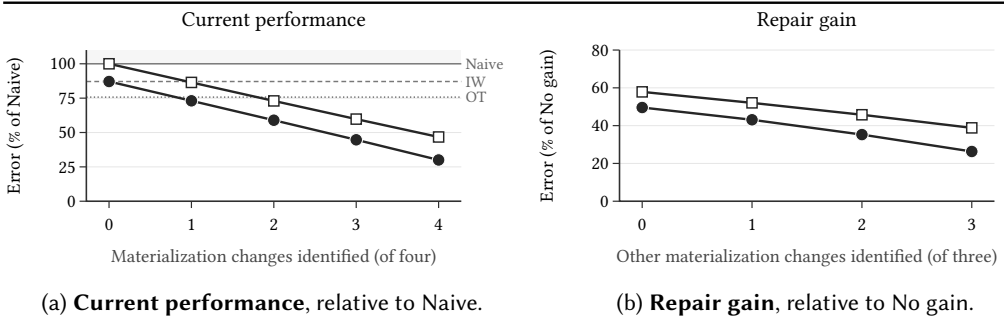
\EOFinish

So far, we assumed that every introduced change is transported (Section~\ref{sec:experimental_setup}). In practice, some materialization changes may not be identified or a corresponding transport may not be available. To examine how this affects the estimands in RQ1 and RQ2, we combine population change with four concurrent faults and vary how many are identified, while keeping the full incident as the estimation target. Unidentified faults are not transported, and their affected variables remain available to population weighting. For repair gain, the selected repair is always identified. At each level, we average error over all corresponding fault subsets and report it relative to Naive for current performance and No gain for repair gain.

Figure~\ref{fig:multifault} shows that mean estimation error decreases as more faults are identified. For current performance, $\mathrm{Transport}(G,M)$ has lower error than IW in every dataset-metric combination once one fault is identified, and a lower mean error than OT from one. For repair gain, it has lower error than No gain in every dataset-metric combination, even when none of the other faults is identified. The reduction from adding population transport grows from 13 to 17 percentage points for current performance and from 8 to 13 for repair gain. This is consistent with unidentified materialization changes limiting both reconstruction and population transport.

\section{Illustrative Case: Monitoring a Churn Prediction Model}
\label{sec:case}

We use the churn prediction case to illustrate what online model testing can provide in a monitoring setting when the required mechanism observability is available. We consider a LightGBM model over twelve deployment weeks and introduce five materialization faults on top of the distribution shifts already present in the data~\cite{kkbox}. The case is therefore illustrative rather than an evaluation of an operational deployment. This lets us examine how the resulting performance estimates and repair tests evolve as several monitored mechanisms change together while labels become available only after 37 days.

\input{sections/figures/case_study}

Figure~\ref{fig:case-study} follows age missingness (M1), a failed profile join (M2), activity snapshot staleness (F1), a changed payment mapping (C1), and a workflow update changing monetary units (C2) as they evolve over time. For each batch, a two-sample test (Fisher’s exact test; significance value $0.1$) is used to flag changes in the monitored mechanisms, and flagged mechanisms are transported to estimate current model performance, and repair gains (for weeks 6, 9, and 12).

\noindent\textbf{Performance estimation (Q1).}
The composed estimate reduces average estimation error relative to IW, but underestimates performance in week 4 and misses the sharp deterioration in week 8. The latter is already present without the injected materialization changes and coincides with a near doubling of the observed churn rate from the preceding week. Its underlying cause is not established, but the constructed transports do not reproduce its effect on performance.

\noindent\textbf{Repair prioritisation (Q2).}
The displayed weeks distinguish repair opportunities among concurrent alerts. In week 6, the profile join fails for a substantial share of observations but offers little repair gain, while payment mapping has the largest estimated and measured gain. A substantial change in a monitored mechanism therefore need not correspond to a substantial effect on model performance. In week 9, correcting the monetary units offers by far the largest measured repair gain. The composed estimate identifies this, although it underestimates the benefit, while without population transport the same repair is estimated to have a slightly negative effect. Accounting for population change can therefore change which repair the tests favor. In week 12, correcting the monetary units remains the largest opportunity, while restoring the increasingly stale activity snapshot actually lowers F1 because the increase in precision is offset by lower recall -- restoring a mechanism to its reference state does not always improve performance for a particular metric.

\section{Discussion}
\label{sec:discussion}

\subsection{Discussion of results}

\noindent\textbf{When population and materialization changes occur jointly, performance estimation benefits from reproducing both.} In the first experiment, composing the declared transports generally improves estimation over either transport alone and over the IW and OT baselines (RQ1). These baselines reproduce a single, assumed source of shift, while the evaluated shifts combine population and materialization changes. Across datasets and materialization mechanisms, these results provide empirical support for distinguishing population and materialization changes to improve performance estimation (Section~\ref{sec:problem}), while the materialization transport template supports the different reconstruction operations these mechanisms require (Section~\ref{sec:methodology}).

\noindent\textbf{Reproducing population change improves estimates
of the gain from repairing a materialization change.}
In the second experiment, representing the evaluation population
in both the faulty and repaired tests generally improves repair
gain estimation (RQ2). This provides empirical support for the
interaction between population and materialization changes
described in Section~\ref{sec:problem}. Adding evaluation labels
to population transport can further improve these estimates,
with larger improvements for some metrics as the populations
separate (Figure~\ref{fig:e21-mixtures}). This suggests that
population differences left unrepresented by IW on comparable
data can become more consequential under larger shifts
(Section~\ref{sec:population_transport}).

\noindent\textbf{Partial mechanism observability and transport remain useful for estimation}. Reproducing other materialization changes improves repair gain estimation even without population transport, supporting the argument that these changes can modulate the effect of the selected repair. Population weighting also lowers mean error for both estimates relative to unweighted tests, even when unidentified materialization changes affect the variables used for weighting. Averaged over fault subsets, errors decrease monotonically as more changes are identified and reproduced (Figure~\ref{fig:multifault}). Practically, these results suggest that the tests need not reproduce every materialization change to improve estimation, but benefit from observing and reproducing more of them.

\subsection{Implications}

For research, our formulation suggests viewing performance estimation and causal attribution under distribution shift as forms of online model testing that differ in which changes they reproduce on reference data. Existing methods typically model either an assumed source of shift~\cite{chen2021mandoline,decker2024explanatory} or changes in mechanisms of the observed data~\cite{budhathokik2021why,zhangh2023why}. Our results motivate methods that explicitly distinguish population and materialization changes and their joint effects on model performance.

For practice, the approach builds on metadata and lineage that feature stores and transformation tools already provide (Section~\ref{sec:materialization_transport}). Today, such signals mainly raise data quality alerts that do not show whether an issue affects the model~\cite{shankars2024we,leest2025monitoring}. We therefore believe that integration with these tools is worth exploring, as our methodology naturally builds on the information they already capture. Such integration could reduce the manual effort needed to track mechanism changes and construct tests, and potentially support more informative alerts when monitoring ML deployments.

\subsection{Future Work}
Several extensions could broaden the scope and applicability of our proposed methodology. First, our experiments consider relatively shallow dependencies between source data and model features, while production ML systems may contain deep and wide pipelines in which changes propagate across many operations. Representing these dependencies could support more effective root-cause analysis and causal attribution \cite{leest2025monitoring}, but may also make observability and transport construction more difficult. Second, the failure modes and transports considered here could form the basis for a catalogue of canonical materialization changes across ML systems and application domains. Further work could compare alternative transports for the same failure mode and examine conditional fault assignment (where the probability that a row is affected depends on context)\footnote{The replication package (Section~\ref{sec:data_availability}) includes separate comparisons
of context dependent group faults and probabilistic snapshot
reconstruction on earlier experimental designs. These results
are not pooled with the main evaluation.}. Third, the formulation could be extended to label materialization. Changes in how outcomes are defined or observed can affect measured performance (e.g., a label definition change). Making this mechanism explicit could distinguish changes in model performance from changes in its measurement.

Our evaluation uses real datasets with simulated faults. Future work should evaluate the approach in production settings, for example through case studies or action research with industry partners, to assess whether real incidents can be observed and reproduced with available observability, how much effort and expertise are needed to declare and maintain transports, how accurate and useful the resulting estimates are, and what runtime overhead the approach introduces.

\section{Threats to Validity}
\label{sec:limitations}

\noindent\textbf{Construct validity.}
The (temporal) datasets provide naturally occurring differences between reference and evaluation periods, which we use to represent population change. These differences may also reflect changes in data collection or processing already present in the datasets and should therefore not be interpreted as population change alone. This does not affect the measured incremental effects of the injected materialization changes, which compare the same evaluation conditions with and without the selected change.

\noindent\textbf{Internal validity.}
The experiments assume that the evaluated materialization changes are correctly identified and that the observations required by their transports are available. This isolates RQ1 and RQ2 but creates more favorable conditions than may hold for real incidents. The partial observability experiment (\ref{sec:experiment_partial}) relaxes complete identification, but does not evaluate incorrect identification. Injecting materialization changes allows us to compare paired conditions with and without each change, so their effects on model performance can be measured directly across real-world population shifts (as is established practice in controlled evaluations of data quality and data cleaning~\cite{arocena2015messing,abedjan2016detecting,schelters2021jenga,jager2024data}). Real incidents may have more complex dependencies, so our evaluation tests the methodology conditional on adequate mechanism observations and transport specifications rather than fault discovery or transport construction in operational systems.

\noindent\textbf{External validity.}
We evaluate six materialization changes across multiple real-world tabular datasets, models, and population conditions. These settings cover several forms of missingness, freshness, and consistency, but not the broader range of mechanisms that may occur in production. The evaluated pipelines also contain relatively shallow dependencies between source data and model features, limiting how directly the findings extend to other data modalities, deeply interconnected pipelines, or continually retrained models. The churn case illustrates how the methodology can support monitoring and repair prioritisation under concurrent changes, but remains a controlled illustration rather than evidence of feasibility or effectiveness in an operational deployment.

\section{Conclusion}
\label{sec:conclusion}

This paper formulates online model testing under distribution shift by distinguishing population changes from changes in how model features are materialized. We show how observations of these mechanisms can be translated into transports and composed on labeled reference data to estimate current performance and repair gains. Under the controlled settings studied here, separating and representing population and materialization changes generally improves both estimates, providing empirical support for testing their effects jointly.

\section{Data Availability}\label{sec:data_availability}
We have made the replication package available at \url{https://doi.org/10.6084/m9.figshare.34062558}. It contains the implementation, data preparation and experiment scripts, results, and scripts for generating the figures. Source data for all twelve \textit{TableShift} datasets and \textit{Default} are included. To respect redistribution terms, the datasets for \textit{Churn} and \textit{Trip} are obtained separately from Kaggle, for which the package provides exact download instructions. We retain archived copies of these source files.

% \bibliographystyle{ACM-Reference-Format}
% \bibliography{bibliography}

\begin{thebibliography}{75}

%%% the Web address.

\ifx \showCODEN    \undefined \def \showCODEN     #1{\unskip}     \fi
\ifx \showISBNx    \undefined \def \showISBNx     #1{\unskip}     \fi
\ifx \showISBNxiii \undefined \def \showISBNxiii  #1{\unskip}     \fi
\ifx \showISSN     \undefined \def \showISSN      #1{\unskip}     \fi
\ifx \showLCCN     \undefined \def \showLCCN      #1{\unskip}     \fi
\ifx \shownote     \undefined \def \shownote      #1{#1}          \fi
\ifx \showarticletitle \undefined \def \showarticletitle #1{#1}   \fi
\ifx \showURL      \undefined \def \showURL       {\relax}        \fi
% invisible to TeX
\providecommand\bibfield[2]{#2}
\providecommand\bibinfo[2]{#2}
\providecommand\natexlab[1]{#1}
\providecommand\showeprint[2][]{arXiv:#2}

\bibitem[Abedjan et~al\mbox{.}(2016)]%
        {abedjan2016detecting}
\bibfield{author}{\bibinfo{person}{Ziawasch Abedjan}, \bibinfo{person}{Xu Chu}, \bibinfo{person}{Dong Deng}, \bibinfo{person}{Raul~Castro Fernandez}, \bibinfo{person}{Ihab~F Ilyas}, \bibinfo{person}{Mourad Ouzzani}, \bibinfo{person}{Paolo Papotti}, \bibinfo{person}{Michael Stonebraker}, {and} \bibinfo{person}{Nan Tang}.} \bibinfo{year}{2016}\natexlab{}.
\newblock \showarticletitle{Detecting data errors: Where are we and what needs to be done?}
\newblock \bibinfo{journal}{\emph{Proceedings of the VLDB Endowment}} \bibinfo{volume}{9}, \bibinfo{number}{12} (\bibinfo{year}{2016}), \bibinfo{pages}{993--1004}.
\newblock


\bibitem[Amershi et~al\mbox{.}(2019)]%
        {amershi2019software}
\bibfield{author}{\bibinfo{person}{Saleema Amershi}, \bibinfo{person}{Andrew Begel}, \bibinfo{person}{Christian Bird}, \bibinfo{person}{Robert DeLine}, \bibinfo{person}{Harald Gall}, \bibinfo{person}{Ece Kamar}, \bibinfo{person}{Nachiappan Nagappan}, \bibinfo{person}{Besmira Nushi}, {and} \bibinfo{person}{Thomas Zimmermann}.} \bibinfo{year}{2019}\natexlab{}.
\newblock \showarticletitle{Software engineering for machine learning: A case study}. In \bibinfo{booktitle}{\emph{2019 IEEE/ACM 41st International Conference on Software Engineering: Software Engineering in Practice (ICSE-SEIP)}}. IEEE, \bibinfo{pages}{291--300}.
\newblock


\bibitem[Ariza-Garz{\'o}n et~al\mbox{.}(2020)]%
        {ariza2020explainability}
\bibfield{author}{\bibinfo{person}{Miller~Janny Ariza-Garz{\'o}n}, \bibinfo{person}{Javier Arroyo}, \bibinfo{person}{Antonio Caparrini}, {and} \bibinfo{person}{Maria-Jesus Segovia-Vargas}.} \bibinfo{year}{2020}\natexlab{}.
\newblock \showarticletitle{Explainability of a machine learning granting scoring model in peer-to-peer lending}.
\newblock \bibinfo{journal}{\emph{Ieee Access}}  \bibinfo{volume}{8} (\bibinfo{year}{2020}), \bibinfo{pages}{64873--64890}.
\newblock


\bibitem[Arocena et~al\mbox{.}(2015)]%
        {arocena2015messing}
\bibfield{author}{\bibinfo{person}{Patricia~C Arocena}, \bibinfo{person}{Boris Glavic}, \bibinfo{person}{Giansalvatore Mecca}, \bibinfo{person}{Ren{\'e}e~J Miller}, \bibinfo{person}{Paolo Papotti}, {and} \bibinfo{person}{Donatello Santoro}.} \bibinfo{year}{2015}\natexlab{}.
\newblock \showarticletitle{Messing up with BART: error generation for evaluating data-cleaning algorithms}.
\newblock \bibinfo{journal}{\emph{Proceedings of the VLDB Endowment}} \bibinfo{volume}{9}, \bibinfo{number}{2} (\bibinfo{year}{2015}), \bibinfo{pages}{36--47}.
\newblock


\bibitem[Auer et~al\mbox{.}(2021)]%
        {auer2021controlled}
\bibfield{author}{\bibinfo{person}{Florian Auer}, \bibinfo{person}{Rasmus Ros}, \bibinfo{person}{Lukas Kaltenbrunner}, \bibinfo{person}{Per Runeson}, {and} \bibinfo{person}{Michael Felderer}.} \bibinfo{year}{2021}\natexlab{}.
\newblock \showarticletitle{Controlled experimentation in continuous experimentation: Knowledge and challenges}.
\newblock \bibinfo{journal}{\emph{Information and Software Technology}}  \bibinfo{volume}{134} (\bibinfo{year}{2021}), \bibinfo{pages}{106551}.
\newblock


\bibitem[Batini et~al\mbox{.}(2009)]%
        {batini2009methodologies}
\bibfield{author}{\bibinfo{person}{Carlo Batini}, \bibinfo{person}{Cinzia Cappiello}, \bibinfo{person}{Chiara Francalanci}, {and} \bibinfo{person}{Andrea Maurino}.} \bibinfo{year}{2009}\natexlab{}.
\newblock \showarticletitle{Methodologies for data quality assessment and improvement}.
\newblock \bibinfo{journal}{\emph{ACM computing surveys (CSUR)}} \bibinfo{volume}{41}, \bibinfo{number}{3} (\bibinfo{year}{2009}), \bibinfo{pages}{1--52}.
\newblock


\bibitem[Bia{\l}ek et~al\mbox{.}(2026)]%
        {bialek2026estimating}
\bibfield{author}{\bibinfo{person}{Jakub Bia{\l}ek}, \bibinfo{person}{Juhani Kivim{\"a}ki}, \bibinfo{person}{Wojciech Kuberski}, {and} \bibinfo{person}{Nikolaos Perrakis}.} \bibinfo{year}{2026}\natexlab{}.
\newblock \showarticletitle{Estimating model performance under covariate shift without labels}.
\newblock \bibinfo{journal}{\emph{Advances in Neural Information Processing Systems}}  \bibinfo{volume}{38} (\bibinfo{year}{2026}), \bibinfo{pages}{161084--161115}.
\newblock


\bibitem[Bifet and Gavald{\`a}(2007)]%
        {bifet2007learning}
\bibfield{author}{\bibinfo{person}{Albert Bifet} {and} \bibinfo{person}{Ricard Gavald{\`a}}.} \bibinfo{year}{2007}\natexlab{}.
\newblock \showarticletitle{Learning from Time-Changing Data with Adaptive Windowing}. In \bibinfo{booktitle}{\emph{Proceedings of the 2007 SIAM International Conference on Data Mining}}. \bibinfo{publisher}{Society for Industrial and Applied Mathematics}, \bibinfo{pages}{443--448}.
\newblock
\href{https://doi.org/10.1137/1.9781611972771.42}{doi:\nolinkurl{10.1137/1.9781611972771.42}}


\bibitem[Bohannon et~al\mbox{.}(2007)]%
        {bohannon2007conditional}
\bibfield{author}{\bibinfo{person}{Philip Bohannon}, \bibinfo{person}{Wenfei Fan}, \bibinfo{person}{Floris Geerts}, \bibinfo{person}{Xibei Jia}, {and} \bibinfo{person}{Anastasios Kementsietsidis}.} \bibinfo{year}{2007}\natexlab{}.
\newblock \showarticletitle{Conditional functional dependencies for data cleaning}. In \bibinfo{booktitle}{\emph{2007 IEEE 23rd international conference on data engineering}}. IEEE, \bibinfo{pages}{746--755}.
\newblock


\bibitem[Bou{\'e} et~al\mbox{.}(2023)]%
        {heynhm2023automotive}
\bibfield{author}{\bibinfo{person}{Laurent Bou{\'e}}, \bibinfo{person}{Pratap Kunireddy}, {and} \bibinfo{person}{Pavle Suboti{\'c}}.} \bibinfo{year}{2023}\natexlab{}.
\newblock \showarticletitle{Automatically Resolving Data Source Dependency Hell in Large Scale Data Science Projects}. In \bibinfo{booktitle}{\emph{2023 IEEE/ACM 2nd International Conference on AI Engineering--Software Engineering for AI (CAIN)}}. IEEE.
\newblock


\bibitem[Breck et~al\mbox{.}(2017)]%
        {breck2017ml}
\bibfield{author}{\bibinfo{person}{Eric Breck}, \bibinfo{person}{Shanqing Cai}, \bibinfo{person}{Eric Nielsen}, \bibinfo{person}{Michael Salib}, {and} \bibinfo{person}{D Sculley}.} \bibinfo{year}{2017}\natexlab{}.
\newblock \showarticletitle{The ML test score: A rubric for ML production readiness and technical debt reduction}. In \bibinfo{booktitle}{\emph{2017 IEEE International Conference on Big Data (Big Data)}}. IEEE, \bibinfo{pages}{1123--1132}.
\newblock


\bibitem[Breck et~al\mbox{.}(2019)]%
        {breck2019data}
\bibfield{author}{\bibinfo{person}{Eric Breck}, \bibinfo{person}{Neoklis Polyzotis}, \bibinfo{person}{Sudip Roy}, \bibinfo{person}{Steven Whang}, {and} \bibinfo{person}{Martin Zinkevich}.} \bibinfo{year}{2019}\natexlab{}.
\newblock \showarticletitle{Data Validation for Machine Learning.}. In \bibinfo{booktitle}{\emph{MLSys}}.
\newblock


\bibitem[Budhathoki et~al\mbox{.}(2021)]%
        {budhathokik2021why}
\bibfield{author}{\bibinfo{person}{Kailash Budhathoki}, \bibinfo{person}{Dominik Janzing}, \bibinfo{person}{Patrick Bl{\"o}baum}, {and} \bibinfo{person}{Hoiyi Ng}.} \bibinfo{year}{2021}\natexlab{}.
\newblock \showarticletitle{Why did the distribution change?}. In \bibinfo{booktitle}{\emph{Proceedings of Machine Learning Research}}.
\newblock


\bibitem[Cai et~al\mbox{.}(2026)]%
        {cai2026diagnosing}
\bibfield{author}{\bibinfo{person}{Tiffany Cai}, \bibinfo{person}{Hongseok Namkoong}, {and} \bibinfo{person}{Steve Yadlowsky}.} \bibinfo{year}{2026}\natexlab{}.
\newblock \showarticletitle{Diagnosing model performance under distribution shift}.
\newblock \bibinfo{journal}{\emph{Operations Research}} \bibinfo{volume}{74}, \bibinfo{number}{2} (\bibinfo{year}{2026}), \bibinfo{pages}{898--916}.
\newblock


\bibitem[Caveness et~al\mbox{.}(2020)]%
        {cavenesse2020tensorflow}
\bibfield{author}{\bibinfo{person}{Emily Caveness}, \bibinfo{person}{Paul~Suganthan GC}, \bibinfo{person}{Zhuo Peng}, \bibinfo{person}{Neoklis Polyzotis}, \bibinfo{person}{Sudip Roy}, {and} \bibinfo{person}{Martin Zinkevich}.} \bibinfo{year}{2020}\natexlab{}.
\newblock \showarticletitle{Tensorflow data validation: Data analysis and validation in continuous ml pipelines}. In \bibinfo{booktitle}{\emph{Proceedings of the 2020 ACM SIGMOD International Conference on Management of Data}}.
\newblock


\bibitem[Chen et~al\mbox{.}(2021)]%
        {chen2021mandoline}
\bibfield{author}{\bibinfo{person}{Mayee Chen}, \bibinfo{person}{Karan Goel}, \bibinfo{person}{Nimit~S Sohoni}, \bibinfo{person}{Fait Poms}, \bibinfo{person}{Kayvon Fatahalian}, {and} \bibinfo{person}{Christopher R{\'e}}.} \bibinfo{year}{2021}\natexlab{}.
\newblock \showarticletitle{Mandoline: Model evaluation under distribution shift}. In \bibinfo{booktitle}{\emph{International conference on machine learning}}. PMLR, \bibinfo{pages}{1617--1629}.
\newblock


\bibitem[Chu et~al\mbox{.}(2013)]%
        {chu2013discovering}
\bibfield{author}{\bibinfo{person}{Xu Chu}, \bibinfo{person}{Ihab~F Ilyas}, {and} \bibinfo{person}{Paolo Papotti}.} \bibinfo{year}{2013}\natexlab{}.
\newblock \showarticletitle{Discovering Denial Constraints.}
\newblock \bibinfo{journal}{\emph{Proc. VLDB Endow.}} \bibinfo{volume}{6}, \bibinfo{number}{13} (\bibinfo{year}{2013}), \bibinfo{pages}{1498--1509}.
\newblock


\bibitem[Dal~Pozzolo et~al\mbox{.}(2017)]%
        {dal2017credit}
\bibfield{author}{\bibinfo{person}{Andrea Dal~Pozzolo}, \bibinfo{person}{Giacomo Boracchi}, \bibinfo{person}{Olivier Caelen}, \bibinfo{person}{Cesare Alippi}, {and} \bibinfo{person}{Gianluca Bontempi}.} \bibinfo{year}{2017}\natexlab{}.
\newblock \showarticletitle{Credit card fraud detection: a realistic modeling and a novel learning strategy}.
\newblock \bibinfo{journal}{\emph{IEEE transactions on neural networks and learning systems}} \bibinfo{volume}{29}, \bibinfo{number}{8} (\bibinfo{year}{2017}), \bibinfo{pages}{3784--3797}.
\newblock


\bibitem[{dbt Labs}(2026)]%
        {dbt_sources}
\bibfield{author}{\bibinfo{person}{{dbt Labs}}.} \bibinfo{year}{2026}\natexlab{}.
\newblock \bibinfo{title}{Add sources to your {DAG}}.
\newblock \bibinfo{howpublished}{\url{https://docs.getdbt.com/docs/build/sources}}.
\newblock


\bibitem[de~la R{\'u}a~Mart{\'\i}nez et~al\mbox{.}(2024)]%
        {de2024hopsworks}
\bibfield{author}{\bibinfo{person}{Javier de~la R{\'u}a~Mart{\'\i}nez}, \bibinfo{person}{Fabio Buso}, \bibinfo{person}{Antonios Kouzoupis}, \bibinfo{person}{Alexandru~A Ormenisan}, \bibinfo{person}{Salman Niazi}, \bibinfo{person}{Davit Bzhalava}, \bibinfo{person}{Kenneth Mak}, \bibinfo{person}{Victor Jouffrey}, \bibinfo{person}{Mikael Ronstr{\"o}m}, \bibinfo{person}{Raymond Cunningham}, {et~al\mbox{.}}} \bibinfo{year}{2024}\natexlab{}.
\newblock \showarticletitle{The hopsworks feature store for machine learning}. In \bibinfo{booktitle}{\emph{Companion of the 2024 International Conference on Management of Data}}. \bibinfo{pages}{135--147}.
\newblock


\bibitem[Decker et~al\mbox{.}(2024)]%
        {decker2024explanatory}
\bibfield{author}{\bibinfo{person}{Thomas Decker}, \bibinfo{person}{Alexander Koebler}, \bibinfo{person}{Michael Lebacher}, \bibinfo{person}{Ingo Thon}, \bibinfo{person}{Volker Tresp}, {and} \bibinfo{person}{Florian Buettner}.} \bibinfo{year}{2024}\natexlab{}.
\newblock \showarticletitle{Explanatory model monitoring to understand the effects of feature shifts on performance}. In \bibinfo{booktitle}{\emph{Proceedings of the 30th ACM SIGKDD Conference on Knowledge Discovery and Data Mining}}. \bibinfo{pages}{550--561}.
\newblock


\bibitem[{gaborfodor}({[n.\,d.]})]%
        {taxi_Kaggle_solution}
\bibfield{author}{\bibinfo{person}{{gaborfodor}}.} \bibinfo{year}{[n.\,d.]}\natexlab{}.
\newblock \bibinfo{title}{From EDA to the Top (LB 0.367)}.
\newblock
\urldef\tempurl%
\url{https://www.kaggle.com/code/gaborfodor/from-eda-to-the-top-lb-0-367}
\showURL{%
\tempurl}


\bibitem[Gama et~al\mbox{.}(2014)]%
        {gama2014survey}
\bibfield{author}{\bibinfo{person}{Jo{\~a}o Gama}, \bibinfo{person}{Indr{\.e} {\v{Z}}liobait{\.e}}, \bibinfo{person}{Albert Bifet}, \bibinfo{person}{Mykola Pechenizkiy}, {and} \bibinfo{person}{Abdelhamid Bouchachia}.} \bibinfo{year}{2014}\natexlab{}.
\newblock \showarticletitle{A survey on concept drift adaptation}.
\newblock \bibinfo{journal}{\emph{ACM computing surveys (CSUR)}} \bibinfo{volume}{46}, \bibinfo{number}{4} (\bibinfo{year}{2014}), \bibinfo{pages}{1--37}.
\newblock


\bibitem[Gardner et~al\mbox{.}(2023)]%
        {gardner2023tableshift}
\bibfield{author}{\bibinfo{person}{Josh Gardner}, \bibinfo{person}{Zoran Popovic}, {and} \bibinfo{person}{Ludwig Schmidt}.} \bibinfo{year}{2023}\natexlab{}.
\newblock \showarticletitle{Benchmarking distribution shift in tabular data with tableshift}.
\newblock \bibinfo{journal}{\emph{Advances in Neural Information Processing Systems}}  \bibinfo{volume}{36} (\bibinfo{year}{2023}), \bibinfo{pages}{53385--53432}.
\newblock


\bibitem[Garg et~al\mbox{.}(2022)]%
        {garg2022leveraging}
\bibfield{author}{\bibinfo{person}{Saurabh Garg}, \bibinfo{person}{Sivaraman Balakrishnan}, \bibinfo{person}{Zachary~C Lipton}, \bibinfo{person}{Behnam Neyshabur}, {and} \bibinfo{person}{Hanie Sedghi}.} \bibinfo{year}{2022}\natexlab{}.
\newblock \showarticletitle{Leveraging unlabeled data to predict out-of-distribution performance}.
\newblock \bibinfo{journal}{\emph{arXiv preprint arXiv:2201.04234}} (\bibinfo{year}{2022}).
\newblock


\bibitem[Grafberger et~al\mbox{.}(2023)]%
        {grafberger2023automating}
\bibfield{author}{\bibinfo{person}{Stefan Grafberger}, \bibinfo{person}{Paul Groth}, {and} \bibinfo{person}{Sebastian Schelter}.} \bibinfo{year}{2023}\natexlab{}.
\newblock \showarticletitle{Automating and optimizing data-centric what-if analyses on native machine learning pipelines}.
\newblock \bibinfo{journal}{\emph{Proceedings of the ACM on Management of Data}} \bibinfo{volume}{1}, \bibinfo{number}{2} (\bibinfo{year}{2023}), \bibinfo{pages}{1--26}.
\newblock


\bibitem[Huyen(2022)]%
        {huyen2022}
\bibfield{author}{\bibinfo{person}{Chip Huyen}.} \bibinfo{year}{2022}\natexlab{}.
\newblock \bibinfo{booktitle}{\emph{Designing machine learning systems}}.
\newblock \bibinfo{publisher}{" O'Reilly Media, Inc."}.
\newblock


\bibitem[{infinitewing}({[n.\,d.]})]%
        {infinitewing_kkbox_solution}
\bibfield{author}{\bibinfo{person}{{infinitewing}}.} \bibinfo{year}{[n.\,d.]}\natexlab{}.
\newblock \bibinfo{title}{Kaggle WSDM--KKBOX's Churn Prediction Challenge Solution sharing}.
\newblock
\urldef\tempurl%
\url{https://www.kaggle.com/competitions/kkbox-churn-prediction-challenge/writeups/infinitewing-solution-sharing}
\showURL{%
\tempurl}


\bibitem[J{\"a}ger and Biessmann(2024)]%
        {jager2024data}
\bibfield{author}{\bibinfo{person}{Sebastian J{\"a}ger} {and} \bibinfo{person}{Felix Biessmann}.} \bibinfo{year}{2024}\natexlab{}.
\newblock \showarticletitle{From data imputation to data cleaning—automated cleaning of tabular data improves downstream predictive performance}. In \bibinfo{booktitle}{\emph{International Conference on Artificial Intelligence and Statistics}}. PMLR, \bibinfo{pages}{3394--3402}.
\newblock


\bibitem[Jung et~al\mbox{.}(2025)]%
        {jung2025towards}
\bibfield{author}{\bibinfo{person}{Philipp Jung}, \bibinfo{person}{Sebastian J{\"a}ger}, \bibinfo{person}{Nicholas Chandler}, {and} \bibinfo{person}{Felix Biessmann}.} \bibinfo{year}{2025}\natexlab{}.
\newblock \showarticletitle{Towards realistic error models for tabular data}.
\newblock \bibinfo{journal}{\emph{ACM Journal of Data and Information Quality}} \bibinfo{volume}{17}, \bibinfo{number}{4} (\bibinfo{year}{2025}), \bibinfo{pages}{1--27}.
\newblock


\bibitem[{Kaggle}({[n.\,d.]})]%
        {taxi_Kaggle}
\bibfield{author}{\bibinfo{person}{{Kaggle}}.} \bibinfo{year}{[n.\,d.]}\natexlab{}.
\newblock \bibinfo{title}{New York City Taxi Trip Duration}.
\newblock
\urldef\tempurl%
\url{https://www.kaggle.com/competitions/nyc-taxi-trip-duration}
\showURL{%
\tempurl}


\bibitem[{KKBOX}({[n.\,d.]})]%
        {kkbox}
\bibfield{author}{\bibinfo{person}{{KKBOX}}.} \bibinfo{year}{[n.\,d.]}\natexlab{}.
\newblock \bibinfo{title}{Kaggle WSDM--KKBOX's Churn Prediction Challenge}.
\newblock
\urldef\tempurl%
\url{https://www.kaggle.com/c/kkbox-churn-prediction-challenge}
\showURL{%
\tempurl}


\bibitem[Kohavi et~al\mbox{.}(2020)]%
        {kohavi2020trustworthy}
\bibfield{author}{\bibinfo{person}{Ron Kohavi}, \bibinfo{person}{Diane Tang}, {and} \bibinfo{person}{Ya Xu}.} \bibinfo{year}{2020}\natexlab{}.
\newblock \bibinfo{booktitle}{\emph{Trustworthy online controlled experiments: A practical guide to a/b testing}}.
\newblock \bibinfo{publisher}{Cambridge University Press}.
\newblock


\bibitem[Kreuzberger et~al\mbox{.}(2023)]%
        {kreuzberger2023machine}
\bibfield{author}{\bibinfo{person}{Dominik Kreuzberger}, \bibinfo{person}{Niklas K{\"u}hl}, {and} \bibinfo{person}{Sebastian Hirschl}.} \bibinfo{year}{2023}\natexlab{}.
\newblock \showarticletitle{Machine learning operations (mlops): Overview, definition, and architecture}.
\newblock \bibinfo{journal}{\emph{IEEE access}}  \bibinfo{volume}{11} (\bibinfo{year}{2023}), \bibinfo{pages}{31866--31879}.
\newblock


\bibitem[Kullback and Leibler(1951)]%
        {kullback1951information}
\bibfield{author}{\bibinfo{person}{Solomon Kullback} {and} \bibinfo{person}{Richard~A Leibler}.} \bibinfo{year}{1951}\natexlab{}.
\newblock \showarticletitle{On information and sufficiency}.
\newblock \bibinfo{journal}{\emph{The annals of mathematical statistics}} \bibinfo{volume}{22}, \bibinfo{number}{1} (\bibinfo{year}{1951}), \bibinfo{pages}{79--86}.
\newblock


\bibitem[Kuroki and Pearl(2014)]%
        {kuroki2014measurement}
\bibfield{author}{\bibinfo{person}{Manabu Kuroki} {and} \bibinfo{person}{Judea Pearl}.} \bibinfo{year}{2014}\natexlab{}.
\newblock \showarticletitle{Measurement bias and effect restoration in causal inference}.
\newblock \bibinfo{journal}{\emph{Biometrika}} (\bibinfo{year}{2014}), \bibinfo{pages}{423--437}.
\newblock


\bibitem[Leest et~al\mbox{.}(2025)]%
        {leest2025monitoring}
\bibfield{author}{\bibinfo{person}{Joran Leest}, \bibinfo{person}{Ilias Gerostathopoulos}, \bibinfo{person}{Patricia Lago}, {and} \bibinfo{person}{Claudia Raibulet}.} \bibinfo{year}{2025}\natexlab{}.
\newblock \showarticletitle{Monitoring and Observability of Machine Learning Systems: Current Practices and Gaps}.
\newblock \bibinfo{journal}{\emph{arXiv preprint arXiv:2510.24142}} (\bibinfo{year}{2025}).
\newblock


\bibitem[Leest et~al\mbox{.}(2026)]%
        {leest2026explaining}
\bibfield{author}{\bibinfo{person}{Joran Leest}, \bibinfo{person}{Ilias Gerostathopoulos}, \bibinfo{person}{Claudia Raibulet}, {and} \bibinfo{person}{Patricia Lago}.} \bibinfo{year}{2026}\natexlab{}.
\newblock \showarticletitle{Explaining Shifts in Machine Learning Systems with Causal Maps}. In \bibinfo{booktitle}{\emph{Proceedings of the IEEE/ACM 5th International Conference on AI Engineering-Software Engineering for AI}}. \bibinfo{pages}{180--185}.
\newblock


\bibitem[Lewis et~al\mbox{.}(2022)]%
        {galewis2022augur}
\bibfield{author}{\bibinfo{person}{GA Lewis}, \bibinfo{person}{S Echeverr{\'i}a}, \bibinfo{person}{L Pons}, {and} \bibinfo{person}{J Chrabaszcz}.} \bibinfo{year}{2022}\natexlab{}.
\newblock \showarticletitle{Augur: A Step Towards Realistic Drift Detection in Production ML Systems}. In \bibinfo{booktitle}{\emph{2022 IEEE/ACM 1st International Workshop on Software Engineering for Responsible Artificial Intelligence (SE4RAI)}}. \bibinfo{publisher}{Association for Computing Machinery}, \bibinfo{pages}{37--44}.
\newblock


\bibitem[Li et~al\mbox{.}(2021)]%
        {li2021cleanml}
\bibfield{author}{\bibinfo{person}{Peng Li}, \bibinfo{person}{Xi Rao}, \bibinfo{person}{Jennifer Blase}, \bibinfo{person}{Yue Zhang}, \bibinfo{person}{Xu Chu}, {and} \bibinfo{person}{Ce Zhang}.} \bibinfo{year}{2021}\natexlab{}.
\newblock \showarticletitle{Cleanml: A study for evaluating the impact of data cleaning on ml classification tasks}. In \bibinfo{booktitle}{\emph{2021 IEEE 37th international conference on data engineering (ICDE)}}. IEEE, \bibinfo{pages}{13--24}.
\newblock


\bibitem[Li et~al\mbox{.}(2017)]%
        {li2017reject}
\bibfield{author}{\bibinfo{person}{Zhiyong Li}, \bibinfo{person}{Ye Tian}, \bibinfo{person}{Ke Li}, \bibinfo{person}{Fanyin Zhou}, {and} \bibinfo{person}{Wei Yang}.} \bibinfo{year}{2017}\natexlab{}.
\newblock \showarticletitle{Reject inference in credit scoring using semi-supervised support vector machines}.
\newblock \bibinfo{journal}{\emph{Expert Systems with Applications}}  \bibinfo{volume}{74} (\bibinfo{year}{2017}), \bibinfo{pages}{105--114}.
\newblock


\bibitem[Lipsitch et~al\mbox{.}(2010)]%
        {lipsitch2010negative}
\bibfield{author}{\bibinfo{person}{Marc Lipsitch}, \bibinfo{person}{Eric~Tchetgen Tchetgen}, {and} \bibinfo{person}{Ted Cohen}.} \bibinfo{year}{2010}\natexlab{}.
\newblock \showarticletitle{Negative controls: a tool for detecting confounding and bias in observational studies}.
\newblock \bibinfo{journal}{\emph{Epidemiology (Cambridge, Mass.)}} \bibinfo{volume}{21}, \bibinfo{number}{3} (\bibinfo{year}{2010}), \bibinfo{pages}{383}.
\newblock


\bibitem[Liu et~al\mbox{.}(2023)]%
        {liu2023need}
\bibfield{author}{\bibinfo{person}{Jiashuo Liu}, \bibinfo{person}{Tianyu Wang}, \bibinfo{person}{Peng Cui}, {and} \bibinfo{person}{Hongseok Namkoong}.} \bibinfo{year}{2023}\natexlab{}.
\newblock \showarticletitle{On the need for a language describing distribution shifts: Illustrations on tabular datasets}.
\newblock \bibinfo{journal}{\emph{Advances in Neural Information Processing Systems}}  \bibinfo{volume}{36} (\bibinfo{year}{2023}), \bibinfo{pages}{51371--51408}.
\newblock


\bibitem[Lu et~al\mbox{.}(2019)]%
        {lu2018}
\bibfield{author}{\bibinfo{person}{Jie Lu}, \bibinfo{person}{Anjin Liu}, \bibinfo{person}{Fan Dong}, \bibinfo{person}{Feng Gu}, \bibinfo{person}{Joao Gama}, {and} \bibinfo{person}{Guangquan Zhang}.} \bibinfo{year}{2019}\natexlab{}.
\newblock \showarticletitle{Learning under concept drift: A review}.
\newblock \bibinfo{journal}{\emph{IEEE Transactions on Knowledge and Data Engineering}} \bibinfo{volume}{31}, \bibinfo{number}{12} (\bibinfo{year}{2019}), \bibinfo{pages}{2346--2363}.
\newblock
\href{https://doi.org/10.1109/TKDE.2018.2876857}{doi:\nolinkurl{10.1109/TKDE.2018.2876857}}


\bibitem[Namaki et~al\mbox{.}(2020)]%
        {namakimh2020vamsa}
\bibfield{author}{\bibinfo{person}{Mohammad~Hossein Namaki}, \bibinfo{person}{Avrilia Floratou}, \bibinfo{person}{Fotis Psallidas}, \bibinfo{person}{Subru Krishnan}, \bibinfo{person}{Ashvin Agrawal}, \bibinfo{person}{Yinghui Wu}, \bibinfo{person}{Yiwen Zhu}, {and} \bibinfo{person}{Markus Weimer}.} \bibinfo{year}{2020}\natexlab{}.
\newblock \showarticletitle{Vamsa: Automated provenance tracking in data science scripts}. In \bibinfo{booktitle}{\emph{Proceedings of the 26th ACM SIGKDD international conference on knowledge discovery \& data mining}}.
\newblock


\bibitem[Naveed et~al\mbox{.}(2025)]%
        {naveed2025monitoring}
\bibfield{author}{\bibinfo{person}{Hira Naveed}, \bibinfo{person}{Scott Barnett}, \bibinfo{person}{Chetan Arora}, \bibinfo{person}{John Grundy}, \bibinfo{person}{Hourieh Khalajzadeh}, {and} \bibinfo{person}{Omar Haggag}.} \bibinfo{year}{2025}\natexlab{}.
\newblock \showarticletitle{Monitoring Machine Learning Systems: A Multivocal Literature Review}.
\newblock \bibinfo{journal}{\emph{arXiv preprint arXiv:2509.14294}} (\bibinfo{year}{2025}).
\newblock


\bibitem[Paleyes et~al\mbox{.}(2023)]%
        {paleyes2023dataflow}
\bibfield{author}{\bibinfo{person}{A Paleyes}, \bibinfo{person}{Siyuan Guo}, \bibinfo{person}{Bernhard Sch{\"o}lkopf}, {and} \bibinfo{person}{ND Lawrence}.} \bibinfo{year}{2023}\natexlab{}.
\newblock \showarticletitle{Dataflow graphs as complete causal graphs}. In \bibinfo{booktitle}{\emph{IEEE/ACM 2nd International Conference on AI Engineering--Software Engineering for AI (CAIN 2023)}}. IEEE, \bibinfo{pages}{7--12}.
\newblock


\bibitem[Papasian and Underwood(2020)]%
        {papasian2020ml}
\bibfield{author}{\bibinfo{person}{Daniel Papasian} {and} \bibinfo{person}{Todd Underwood}.} \bibinfo{year}{2020}\natexlab{}.
\newblock \showarticletitle{How {ML} Breaks: A Decade of Outages for One Large {ML} Pipeline}. \bibinfo{publisher}{USENIX Association}.
\newblock


\bibitem[Papenbrock et~al\mbox{.}(2015)]%
        {papenbrock2015functional}
\bibfield{author}{\bibinfo{person}{Thorsten Papenbrock}, \bibinfo{person}{Jens Ehrlich}, \bibinfo{person}{Jannik Marten}, \bibinfo{person}{Tommy Neubert}, \bibinfo{person}{Jan-Peer Rudolph}, \bibinfo{person}{Martin Sch{\"o}nberg}, \bibinfo{person}{Jakob Zwiener}, {and} \bibinfo{person}{Felix Naumann}.} \bibinfo{year}{2015}\natexlab{}.
\newblock \showarticletitle{Functional dependency discovery: An experimental evaluation of seven algorithms}.
\newblock \bibinfo{journal}{\emph{Proceedings of the VLDB Endowment}} \bibinfo{volume}{8}, \bibinfo{number}{10} (\bibinfo{year}{2015}), \bibinfo{pages}{1082--1093}.
\newblock


\bibitem[Rekatsinas et~al\mbox{.}(2017)]%
        {rekatsinas2017holoclean}
\bibfield{author}{\bibinfo{person}{Theodoros Rekatsinas}, \bibinfo{person}{Xu Chu}, \bibinfo{person}{Ihab~F Ilyas}, {and} \bibinfo{person}{Christopher R{\'e}}.} \bibinfo{year}{2017}\natexlab{}.
\newblock \showarticletitle{Holoclean: Holistic data repairs with probabilistic inference}.
\newblock \bibinfo{journal}{\emph{arXiv preprint arXiv:1702.00820}} (\bibinfo{year}{2017}).
\newblock


\bibitem[Ribeiro et~al\mbox{.}(2020)]%
        {ribeiro2020beyond}
\bibfield{author}{\bibinfo{person}{Marco~Tulio Ribeiro}, \bibinfo{person}{Tongshuang Wu}, \bibinfo{person}{Carlos Guestrin}, {and} \bibinfo{person}{Sameer Singh}.} \bibinfo{year}{2020}\natexlab{}.
\newblock \showarticletitle{Beyond accuracy: Behavioral testing of NLP models with CheckList}. In \bibinfo{booktitle}{\emph{Proceedings of the 58th annual meeting of the association for computational linguistics}}. \bibinfo{pages}{4902--4912}.
\newblock


\bibitem[Riccio et~al\mbox{.}(2020)]%
        {riccio2020testing}
\bibfield{author}{\bibinfo{person}{Vincenzo Riccio}, \bibinfo{person}{Gunel Jahangirova}, \bibinfo{person}{Andrea Stocco}, \bibinfo{person}{Nargiz Humbatova}, \bibinfo{person}{Michael Weiss}, {and} \bibinfo{person}{Paolo Tonella}.} \bibinfo{year}{2020}\natexlab{}.
\newblock \showarticletitle{Testing machine learning based systems: a systematic mapping}.
\newblock \bibinfo{journal}{\emph{Empirical Software Engineering}} \bibinfo{volume}{25}, \bibinfo{number}{6} (\bibinfo{year}{2020}), \bibinfo{pages}{5193--5254}.
\newblock


\bibitem[Runge et~al\mbox{.}(2014)]%
        {runge2014churn}
\bibfield{author}{\bibinfo{person}{Julian Runge}, \bibinfo{person}{Peng Gao}, \bibinfo{person}{Florent Garcin}, {and} \bibinfo{person}{Boi Faltings}.} \bibinfo{year}{2014}\natexlab{}.
\newblock \showarticletitle{Churn prediction for high-value players in casual social games}. In \bibinfo{booktitle}{\emph{2014 IEEE conference on Computational Intelligence and Games}}. IEEE, \bibinfo{pages}{1--8}.
\newblock


\bibitem[Saeed et~al\mbox{.}(2020)]%
        {saeed2020anchored}
\bibfield{author}{\bibinfo{person}{Basil Saeed}, \bibinfo{person}{Anastasiya Belyaeva}, \bibinfo{person}{Yuhao Wang}, {and} \bibinfo{person}{Caroline Uhler}.} \bibinfo{year}{2020}\natexlab{}.
\newblock \showarticletitle{Anchored causal inference in the presence of measurement error}. In \bibinfo{booktitle}{\emph{Conference on uncertainty in artificial intelligence}}. PMLR, \bibinfo{pages}{619--628}.
\newblock


\bibitem[Sanz-Guerrero and Arroyo(2024)]%
        {sanz2024credit}
\bibfield{author}{\bibinfo{person}{Mario Sanz-Guerrero} {and} \bibinfo{person}{Javier Arroyo}.} \bibinfo{year}{2024}\natexlab{}.
\newblock \showarticletitle{Credit risk meets large language models: Building a risk indicator from loan descriptions in p2p lending}.
\newblock \bibinfo{journal}{\emph{arXiv preprint arXiv:2401.16458}} (\bibinfo{year}{2024}).
\newblock


\bibitem[Schelter et~al\mbox{.}(2018)]%
        {schelter2018automating}
\bibfield{author}{\bibinfo{person}{Sebastian Schelter}, \bibinfo{person}{Dustin Lange}, \bibinfo{person}{Philipp Schmidt}, \bibinfo{person}{Meltem Celikel}, \bibinfo{person}{Felix Biessmann}, {and} \bibinfo{person}{Andreas Grafberger}.} \bibinfo{year}{2018}\natexlab{}.
\newblock \showarticletitle{Automating large-scale data quality verification}.
\newblock \bibinfo{journal}{\emph{Proceedings of the VLDB Endowment}} \bibinfo{volume}{11}, \bibinfo{number}{12} (\bibinfo{year}{2018}), \bibinfo{pages}{1781--1794}.
\newblock


\bibitem[Schelter et~al\mbox{.}(2020)]%
        {schelters2020learning}
\bibfield{author}{\bibinfo{person}{Sebastian Schelter}, \bibinfo{person}{Tammo Rukat}, {and} \bibinfo{person}{Felix Bie{\ss}mann}.} \bibinfo{year}{2020}\natexlab{}.
\newblock \showarticletitle{Learning to validate the predictions of black box classifiers on unseen data}. In \bibinfo{booktitle}{\emph{Proceedings of the 2020 ACM SIGMOD International Conference on Management of Data}}.
\newblock


\bibitem[Schelter et~al\mbox{.}(2021)]%
        {schelters2021jenga}
\bibfield{author}{\bibinfo{person}{Sebastian Schelter}, \bibinfo{person}{Tammo Rukat}, {and} \bibinfo{person}{Felix Biessmann}.} \bibinfo{year}{2021}\natexlab{}.
\newblock \showarticletitle{JENGA - A framework to study the impact of data errors on the predictions of machine learning models}.
\newblock \bibinfo{journal}{\emph{Advances in Database Technology - EDBT}} (\bibinfo{year}{2021}).
\newblock


\bibitem[Shankar et~al\mbox{.}(2023)]%
        {shankar2023automatic}
\bibfield{author}{\bibinfo{person}{Shreya Shankar}, \bibinfo{person}{Labib Fawaz}, \bibinfo{person}{Karl Gyllstrom}, {and} \bibinfo{person}{Aditya Parameswaran}.} \bibinfo{year}{2023}\natexlab{}.
\newblock \showarticletitle{Automatic and Precise Data Validation for Machine Learning}. In \bibinfo{booktitle}{\emph{Proceedings of the 32nd ACM International Conference on Information and Knowledge Management}}. \bibinfo{pages}{2198--2207}.
\newblock


\bibitem[Shankar et~al\mbox{.}(2024)]%
        {shankars2024we}
\bibfield{author}{\bibinfo{person}{Shreya Shankar}, \bibinfo{person}{Rolando Garcia}, \bibinfo{person}{Joseph~M Hellerstein}, {and} \bibinfo{person}{Aditya~G Parameswaran}.} \bibinfo{year}{2024}\natexlab{}.
\newblock \showarticletitle{" We Have No Idea How Models will Behave in Production until Production": How Engineers Operationalize Machine Learning}.
\newblock \bibinfo{journal}{\emph{Proceedings of the ACM on Human-Computer Interaction}} \bibinfo{volume}{8}, \bibinfo{number}{CSCW1}, \bibinfo{pages}{1--34}.
\newblock


\bibitem[Shankar and Parameswaran(2022)]%
        {shankars2022towards}
\bibfield{author}{\bibinfo{person}{Shreya Shankar} {and} \bibinfo{person}{Aditya~G Parameswaran}.} \bibinfo{year}{2022}\natexlab{}.
\newblock \showarticletitle{Towards Observability for Machine Learning Pipelines.}. In \bibinfo{booktitle}{\emph{CIDR}}.
\newblock


\bibitem[Sobolewski and Wo{\'z}niak(2017)]%
        {sobolewskip2017scr}
\bibfield{author}{\bibinfo{person}{Piotr Sobolewski} {and} \bibinfo{person}{Micha{\l} Wo{\'z}niak}.} \bibinfo{year}{2017}\natexlab{}.
\newblock \showarticletitle{SCR: simulated concept recurrence--a non-supervised tool for dealing with shifting concept}.
\newblock \bibinfo{journal}{\emph{Expert Systems}} \bibinfo{volume}{34}, \bibinfo{number}{5} (\bibinfo{year}{2017}).
\newblock


\bibitem[Sugiyama et~al\mbox{.}(2007)]%
        {sugiyama2007covariate}
\bibfield{author}{\bibinfo{person}{Masashi Sugiyama}, \bibinfo{person}{Matthias Krauledat}, {and} \bibinfo{person}{Klaus-Robert M{\"u}ller}.} \bibinfo{year}{2007}\natexlab{}.
\newblock \showarticletitle{Covariate shift adaptation by importance weighted cross validation.}
\newblock \bibinfo{journal}{\emph{Journal of Machine Learning Research}} (\bibinfo{year}{2007}).
\newblock


\bibitem[Tecton({[n.\,d.]})]%
        {tecton_materialization_monitoring}
\bibfield{author}{\bibinfo{person}{Tecton}.} \bibinfo{year}{[n.\,d.]}\natexlab{}.
\newblock \bibinfo{title}{Materialization}.
\newblock \bibinfo{howpublished}{Tecton documentation, Monitoring and Alerting}.
\newblock
\urldef\tempurl%
\url{https://docs.tecton.ai/docs/monitoring/materialization}
\showURL{%
\tempurl}


\bibitem[Tecton(2025)]%
        {tecton_http_api}
\bibfield{author}{\bibinfo{person}{Tecton}.} \bibinfo{year}{2025}\natexlab{}.
\newblock \bibinfo{title}{HTTP API}.
\newblock \bibinfo{howpublished}{Tecton documentation, Read Features for Inference}.
\newblock
\urldef\tempurl%
\url{https://docs.tecton.ai/docs/reading-feature-data/reading-feature-data-for-inference/reading-online-features-for-inference-using-the-http-api}
\showURL{%
\tempurl}


\bibitem[{The Feast Authors}(2026a)]%
        {feast_metrics_implementation}
\bibfield{author}{\bibinfo{person}{{The Feast Authors}}.} \bibinfo{year}{2026}\natexlab{a}.
\newblock \bibinfo{title}{Feast feature server metrics implementation}.
\newblock \bibinfo{howpublished}{GitHub source code}.
\newblock
\urldef\tempurl%
\url{https://github.com/feast-dev/feast/blob/104ad10403e4e7dcb27145a86b2523c1c4639612/sdk/python/feast/metrics.py}
\showURL{%
\tempurl}


\bibitem[{The Feast Authors}(2026b)]%
        {feast_feature_view}
\bibfield{author}{\bibinfo{person}{{The Feast Authors}}.} \bibinfo{year}{2026}\natexlab{b}.
\newblock \bibinfo{title}{Feature view}.
\newblock \bibinfo{howpublished}{Feast documentation}.
\newblock
\urldef\tempurl%
\url{https://docs.feast.dev/getting-started/concepts/feature-view}
\showURL{%
\tempurl}


\bibitem[Webb et~al\mbox{.}(2016)]%
        {webb2016characterizing}
\bibfield{author}{\bibinfo{person}{Geoffrey~I Webb}, \bibinfo{person}{Roy Hyde}, \bibinfo{person}{Hong Cao}, \bibinfo{person}{Hai~Long Nguyen}, {and} \bibinfo{person}{Francois Petitjean}.} \bibinfo{year}{2016}\natexlab{}.
\newblock \showarticletitle{Characterizing concept drift}.
\newblock \bibinfo{journal}{\emph{Data Mining and Knowledge Discovery}} \bibinfo{volume}{30}, \bibinfo{number}{4} (\bibinfo{year}{2016}), \bibinfo{pages}{964--994}.
\newblock


\bibitem[Wooders et~al\mbox{.}(2023)]%
        {wooders2023ralf}
\bibfield{author}{\bibinfo{person}{Sarah Wooders}, \bibinfo{person}{Xiangxi Mo}, \bibinfo{person}{Amit Narang}, \bibinfo{person}{Kevin Lin}, \bibinfo{person}{Ion Stoica}, \bibinfo{person}{Joseph~M Hellerstein}, \bibinfo{person}{Natacha Crooks}, {and} \bibinfo{person}{Joseph~E Gonzalez}.} \bibinfo{year}{2023}\natexlab{}.
\newblock \showarticletitle{Ralf: Accuracy-aware scheduling for feature store maintenance}.
\newblock \bibinfo{journal}{\emph{Proceedings of the VLDB Endowment}} \bibinfo{volume}{17}, \bibinfo{number}{3} (\bibinfo{year}{2023}), \bibinfo{pages}{563--576}.
\newblock


\bibitem[Xie et~al\mbox{.}(2011)]%
        {xie2011testing}
\bibfield{author}{\bibinfo{person}{Xiaoyuan Xie}, \bibinfo{person}{Joshua~WK Ho}, \bibinfo{person}{Christian Murphy}, \bibinfo{person}{Gail Kaiser}, \bibinfo{person}{Baowen Xu}, {and} \bibinfo{person}{Tsong~Yueh Chen}.} \bibinfo{year}{2011}\natexlab{}.
\newblock \showarticletitle{Testing and validating machine learning classifiers by metamorphic testing}.
\newblock \bibinfo{journal}{\emph{Journal of Systems and Software}} \bibinfo{volume}{84}, \bibinfo{number}{4} (\bibinfo{year}{2011}), \bibinfo{pages}{544--558}.
\newblock


\bibitem[Zhang et~al\mbox{.}(2023)]%
        {zhangh2023why}
\bibfield{author}{\bibinfo{person}{Haoran Zhang}, \bibinfo{person}{Harvineet Singh}, \bibinfo{person}{Marzyeh Ghassemi}, {and} \bibinfo{person}{Shalmali Joshi}.} \bibinfo{year}{2023}\natexlab{}.
\newblock \showarticletitle{“Why did the Model Fail?”: Attributing Model Performance Changes to Distribution Shifts}. In \bibinfo{booktitle}{\emph{Proceedings of Machine Learning Research}}.
\newblock


\bibitem[Zhang et~al\mbox{.}(2020)]%
        {zhang2020machine}
\bibfield{author}{\bibinfo{person}{Jie~M Zhang}, \bibinfo{person}{Mark Harman}, \bibinfo{person}{Lei Ma}, {and} \bibinfo{person}{Yang Liu}.} \bibinfo{year}{2020}\natexlab{}.
\newblock \showarticletitle{Machine learning testing: Survey, landscapes and horizons}.
\newblock \bibinfo{journal}{\emph{IEEE Transactions on Software Engineering}} \bibinfo{volume}{48}, \bibinfo{number}{1} (\bibinfo{year}{2020}), \bibinfo{pages}{1--36}.
\newblock


\bibitem[Zhou et~al\mbox{.}(2019)]%
        {zhoux2019a}
\bibfield{author}{\bibinfo{person}{Xianzhe Zhou}, \bibinfo{person}{Wally Lo~Faro}, \bibinfo{person}{Xiaoying Zhang}, {and} \bibinfo{person}{Ravi~Santosh Arvapally}.} \bibinfo{year}{2019}\natexlab{}.
\newblock \showarticletitle{A framework to monitor machine learning systems using concept drift detection}.
\newblock  (\bibinfo{year}{2019}), \bibinfo{pages}{218--231}.
\newblock


\bibitem[Zinkevich(2017)]%
        {zinkevich2017rules}
\bibfield{author}{\bibinfo{person}{Martin Zinkevich}.} \bibinfo{year}{2017}\natexlab{}.
\newblock \bibinfo{title}{Rules of Machine Learning: Best Practices for {ML} Engineering}.
\newblock \bibinfo{howpublished}{Google for Developers}.
\newblock
\urldef\tempurl%
\url{https://developers.google.com/machine-learning/guides/rules-of-ml}
\showURL{%
\tempurl}


\bibitem[{\v{Z}}liobait{\.e} et~al\mbox{.}(2015)]%
        {vzliobaite2015overview}
\bibfield{author}{\bibinfo{person}{Indr{\.e} {\v{Z}}liobait{\.e}}, \bibinfo{person}{Mykola Pechenizkiy}, {and} \bibinfo{person}{Joao Gama}.} \bibinfo{year}{2015}\natexlab{}.
\newblock \showarticletitle{An overview of concept drift applications}.
\newblock \bibinfo{journal}{\emph{Big data analysis: new algorithms for a new society}} (\bibinfo{year}{2015}), \bibinfo{pages}{91--114}.
\newblock


\end{thebibliography}

\end{document}